\documentclass{article}
\usepackage{ijcai22}

\usepackage{times}

\usepackage{soul}
\usepackage{url}
\usepackage[hidelinks]{hyperref}
\usepackage[utf8]{inputenc}
\usepackage[small]{caption}
\usepackage{graphicx}
\usepackage{amsmath}
\usepackage{booktabs}
\usepackage{multirow}
\usepackage{subcaption}
\usepackage{enumitem}
\usepackage{amsfonts}

\title{Spatio-Temporal meets Wavelet: Disentangled Traffic Flow Forecasting via Efficient Spectral Graph Attention Network}

\author{
Yuchen Fang$^1$\and
Yanjun Qin$^1$\footnote{Equal contribution.}\and
Haiyong Luo$^{2}$\footnote{Corresponding author.}\and
Fang Zhao$^{1\dag}$\and\\
Bingbing Xu$^3$\and
Chenxing Wang$^1$\And
Liang Zeng$^4$\\
\affiliations
$^1$School of Computer Science (National Pilot Software Engineering School), \\Beijing University of Posts and Telecommunications, China\\
$^2$Research Center for Ubiquitous Computing Systems, \\Institute of Computing Technology, Chinese Academy of Sciences, China\\
$^3$CAS Key Laboratory of Network Data Science and Technology, \\Institute of Computing Technology, Chinese Academy of Sciences, China\\
$^4$Institute for Interdisciplinary Information Sciences (IIIS), Tsinghua University, China\\
\emails
\{fangyuchen, qinyanjun, zfsse, wangchenxing\}@bupt.edu.cn,
\{yhluo,xubingbing\}@ict.ac.cn,
zengl18@mails.tsinghua.edu.cn
}

\begin{document}

\maketitle

\begin{abstract}
Traffic forecasting is crucial for public safety and resource optimization, yet is very challenging due to three aspects: i) current existing works mostly exploit intricate temporal patterns (\emph{e.g.}, the short-term thunderstorm and long-term daily trends) within a single method, which fail to accurately capture spatio-temporal dependencies under different schemas; ii) the under-exploration of the graph positional encoding limit the extraction of spatial information in the commonly used full graph attention network; iii) the quadratic complexity of the full graph attention introduces heavy computational needs. To achieve the effective traffic flow forecasting, we propose an efficient spectral graph attention network with disentangled traffic sequences. Specifically, the discrete wavelet transform is leveraged to obtain the low- and high-frequency components of traffic sequences, and a dual-channel encoder is elaborately designed to accurately capture the spatio-temporal dependencies under long- and short-term schemas of the low- and high-frequency components. Moreover, a novel wavelet-based graph positional encoding and a query sampling strategy are introduced in our spectral graph attention to effectively guide message passing and efficiently calculate the attention. Extensive experiments on four real-world datasets show the superiority of our model, \emph{i.e.}, the higher traffic forecasting precision with lower computational cost.


\end{abstract}

\section{Introduction}
Given the observed traffic conditions and underlying road networks, traffic flow forecasting aims to predict a sequence of future traffic flow, which benefits both daily travel and traffic management. Despite its importance, traffic forecasting is very challenging because of the intricate spatio-temporal dependencies. Recently, data-driven algorithms have received significant attention in the community. Among them, recurrent neural network (RNN), temporal convolution network (TCN), and Transformer-based methods have been widely adopted to capture temporal dependencies for each road individually \cite{zhou2021informer,xu2021autoformer,sutskever2014sequence,lv2018lc,elmi2020deep}. Though the uni-variate model endowed above methods to predict traffic flow, they exploit the entangled temporal patterns within a single model (\emph{e.g.}, using a sequential model to extract the contrary influence of increasing and decreasing flow in car accidents and epidemic), which fail to accurately extract the spatio-temporal information under different schemas.

The adoption of graph convolutional networks (GCNs) \cite{defferrard2016convolutional} has recently solved the issue of capturing spatial correlations in traffic forecasting. DCRNN \cite{DBLP:conf/iclr/LiYS018} and STGCN \cite{DBLP:conf/ijcai/YuYZ18} use GCN to model the interactions of neighboring roads. Subsequent works, such as Graph WaveNet, AGCRN, STFGNN, and STGODE \cite{DBLP:conf/ijcai/WuPLJZ19,DBLP:conf/nips/0001YL0020,DBLP:conf/aaai/LiZ21,fang2021spatial}, propose multiple variants of GCN to improve performance by expanding spatial receptive fields. However, GCN-based methods ignore that the weight of edges in the graph of road networks is constantly changing over time. ST-CGA \cite{zhang2020spatial} and LSGCN \cite{huang2020lsgcn} use the graph attention network (GAT) and its variant to learn the weights between neighbor roads in each time step. ST-GRAT \cite{park2020st} and GMAN \cite{zheng2020gman} further utilize the full GAT for traffic speed forecasting, which drops the input graph in vanilla GAT to alleviate the influence of hard inductive bias and capture the global spatial dependence. Although the full GAT-based works have shown promising performance in other traffic forecasting tasks, most of them suffer from two limitations: 1) they neglect the learning efficiency on the full attention, \emph{i.e.}, the time and space complexity of well-known self-attention is $O(N^2)$, which introduces heavy computational needs; 2) the central issue of vanilla GAT is that the input graph restricts the spatial receptive field into neighbors, and full GAT lacks structural information to effectively guide message passing.


To alleviate the above limitations, we propose a novel efficient spectral graph attention network for traffic flow forecasting with disentangled traffic sequences. For the temporal dimension, to model the intricate temporal patterns individually, we firstly utilize discrete wavelet transform to obtain the low- and high-frequency components from traffic sequences, where the low- and high-frequency components can reflect the temporal patterns of the long- and short-term schemas. Then we carefully design a dual-channel spatio-temporal encoder to represent the different dual-scale temporal patterns. Moreover, we propose a fusion attention to aggregate the latent representations of the dual-scale temporal patterns, and perform multi-supervision to predict a sequence of future low-frequency component, which is in parallel with the traffic flow forecasting. For the spatial dimension, to improve the performance of the full GAT for traffic flow forecasting, we propose an efficient spectral graph attention network, which is a variant of the self-attention with only $O(NlogN)$ complexity. The novel wavelet-based graph positional encoding and query sampling strategy are introduced in our attention to guide message passing and sparsify the nodes in the query of self-attention. The key technical contribution of this paper is a elaborately designed model, namely STWave, which leverages the graph and discrete wavelet on spatio-temporal dimensions simultaneously. Experimental results on four real-world datasets show STWave significantly outperforms state-of-the-arts on traffic flow forecasting.
\section{Preliminaries}
\paragraph{Traffic Network.} Traffic network is defined as an undirected graph $\mathcal{G}=(V,E,A)$, where $V$ is the set of nodes, $E$ is the set of edges between neighboring nodes, and $A\in\mathbb{R}^{N\times N}$ corresponds to the adjacency matrix of $\mathcal{G}$. In practice, a node may represent a sensor located at the corresponding road of the traffic network. Each node records traffic flow. Here in this paper, two kinds of the graph are adopted, $A^{spa}$ is the adjacency matrix of the spatial graph according to the traffic network and $A^{tem}$ is the adjacency matrix of the temporal graph. The weights of $A^{tem}$ derived from the dynamic time warping (DTW) algorithm \cite{berndt1994using} followed by STFGNN \cite{DBLP:conf/aaai/LiZ21}.
\paragraph{Problem Definition.} For a traffic network, let $x_t^i\in\mathbb{R}^C, C=1$ represents the traffic flow value of the $i$th node at time step $t$, and $X_t=[x_t^1,...,x_t^i,...,x_t^N]^T\in\mathbb{R}^{N\times C}$ represents the traffic flow values of all nodes at time step $t$. Given history traffic data $\mathcal{X}=\{X_1,...,X_{T_1}\}\in\mathbb{R}^{T_1\times N\times C}$, the purpose of traffic prediction is to predict the traffic flow of all nodes in future $T_2$ time steps, namely $\mathcal{\hat{Y}}=\{\hat{Y}_{(T_1+1)},...,\hat{Y}_{(T_1+T_2)}\}\in\mathbb{R}^{T_2\times N\times C}$, and its ground truth is denoted by $\mathcal{\hat{X}}=\{X_{(T_1+1)},...,X_{(T_1+T_2)}\}\in\mathbb{R}^{T_2\times N\times C}$.
\paragraph{Self-Attention.} Self-attention is the mostly used attention mechanism. The key idea behind the mechanism is that each element in a sequence learns to gather information from other tokens. The input of self-attention consists of queries, keys, and values of dimension $d$. Then compute the dot products of the query with all keys, divide each by $\sqrt{d}$, and apply a $softmax$ function to obtain the weights on the values:
\begin{small}
\begin{equation}
    Att(Q,K,V)=softmax(\frac{(QW^Q)(K^TW^K)}{\sqrt{d}})(VW^V)\text{  ,}
\end{equation}
\end{small}
where $W^{Q},W^{K},W^{V}$ are learnable parameters of projections, $Att(\cdot)$ denotes the self-attention operation.
\paragraph{Wavelet Transform.} Wavelet \cite{daubechies1992ten} is associated with scaling function and wavelet function, whose shifts and expansions compose stable basis for the signal space. The scaling and wavelet functions of discrete wavelet are closely related with low-pass filter $\mathbf{g}=\{g_k\}_{k\in\mathbb{Z}}$ and high-pass filter $\mathbf{h}=\{h_k\}_{k\in\mathbb{Z}}$, respectively. For a 1D signal $\mathbf{x}=\{x_j\}_{j\in\mathbb{Z}}$, discrete wavelet transform decomposes it into low-frequency component $\mathbf{x_l}=\{x_{l_k}\}_{k\in\mathbb{Z}}$ and high-frequency component $\mathbf{x_h}=\{x_{h_k}\}_{k\in\mathbb{Z}}$:
\begin{equation}
    x_{l_k}=\sum\nolimits_{j}g_{j-2k}x_j\text{, }x_{h_k}=\sum\nolimits_{j}h_{j-2k}x_j\text{  .}\\
\end{equation}
Besides, the wavelet of graph corresponds to the signal on graph diffused away from nodes with a scaling matrix $G_{s}=diag(\sigma(s\lambda_1),...,\sigma(s\lambda_d))$, $\sigma(s\lambda_i)=e^{\lambda_is}$ and $\lambda_i$ is the $i$th lowest graph Laplacian eigenvalues. The expression of the graph wavelet $\psi_s$ at scale $s$ can be formulated as:
\begin{equation}
    \psi_s=\Phi G_s\Phi^T\text{  ,}
\end{equation}
where $\Phi$ is eigenvectors of the graph Laplacian. For a graph signal $x$, the graph wavelet transform is defined as $\tilde{x}=\psi_s^{-1}x$.

\begin{figure*}
    \centering
    \includegraphics[width=1.0\linewidth]{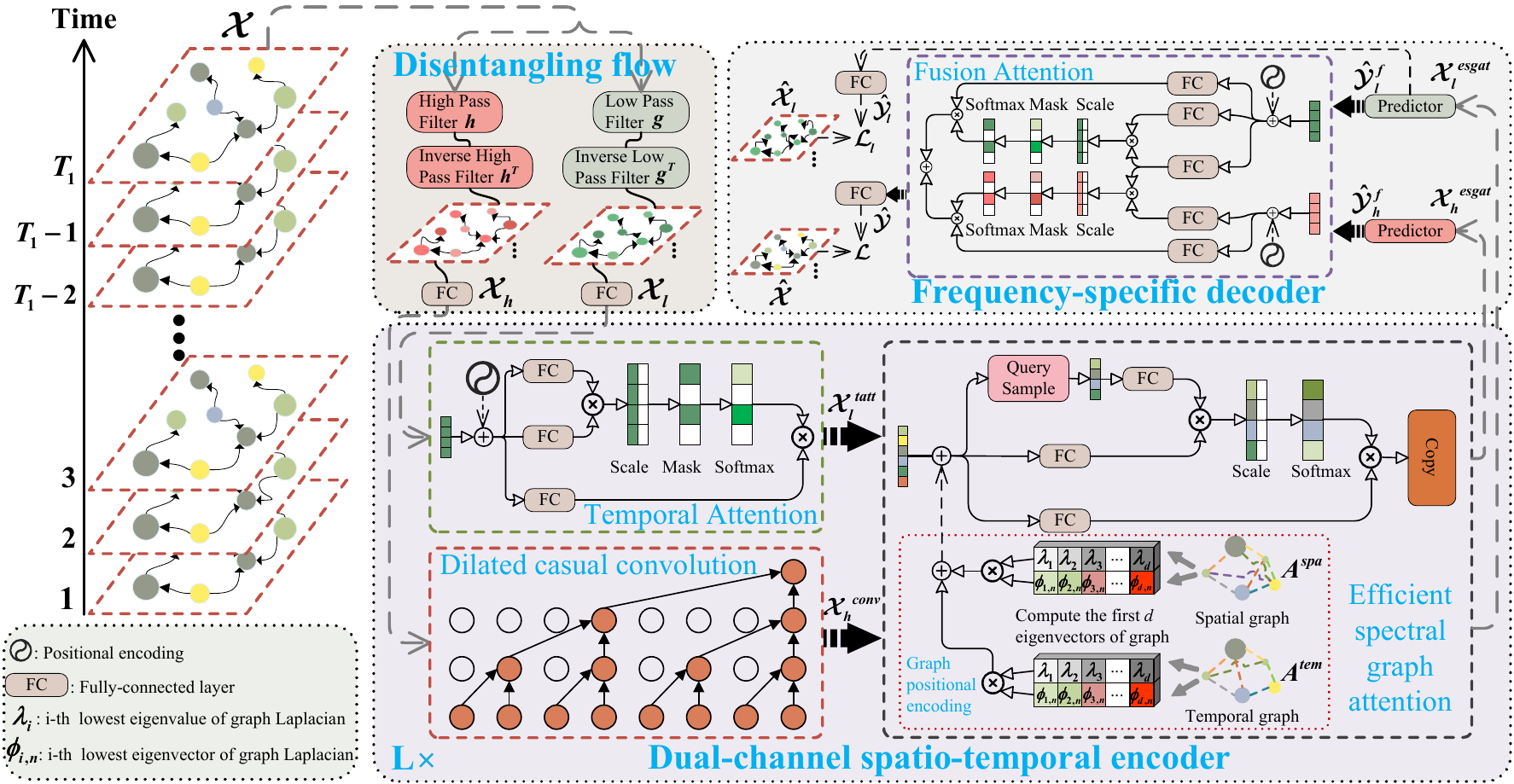}
    \caption{The architecture of the proposed STWave.}
    \label{model}
\end{figure*}

\section{Methodology}
Our STWave is outlined in Figure \ref{model}, consisting of a disentangling flow layer, a dual-channel spatio-temporal encoder, and a frequency-specific decoder. The disentangling flow layer is used to separate entangled long- and short-term temporal patterns to avoid the interference between them. The dual-channel spatio-temporal encoder are stacked by $L$ layers, aims to effectively represent the dual-scale spatio-temporal patterns. Then the fusion attention and multi-supervision are utilized in the frequency-specific decoder to merge and exploit the dual-scale information.
\subsection{Disentangling Flow Layer}
Given the traffic sequences $\mathcal{X}\in\mathbb{R}^{T_1\times N\times d}$, we use the discrete wavelet transform (DWT) to obtain the low- and high-frequency components from the entangled traffic sequences, where the dual-scale components indicate the long- and short-term temporal patterns because the low-frequency component is stable and has long-term trends, while the fluctuate high-frequency has short-term influences. 
The DWT with the input traffic sequences $\mathcal{X}$ can be formulated as:
\begin{equation}
    \mathcal{\Bar{X}}_l=\mathbf{g}\mathcal{X}\text{, }\mathcal{\Bar{X}}_h=\mathbf{h}\mathcal{X}\text{  ,}
\end{equation}
where the time steps in the candidate low- and high-frequency components $\mathcal{\Bar{X}}_l$ and $\mathcal{\Bar{X}}_h$ are reduced to the half of the input by the down-sampling operation in DWT. Therefore, the inverse low- and high-pass filters $\mathbf{g}^T,\mathbf{h}^T$ are adopted in this layer to up-sample to consist with the input. Then we use a fully-connected layer to transform the low- and high-frequency components into high-dimensional low- and high-frequency components $\mathcal{X}_l,\mathcal{X}_h\in\mathbb{R}^{T_1\times N\times d}$, which can improve the representation power of STWave. The up-sampling operation and fully-connected layer are formulated as:
\begin{equation}
    \mathcal{X}_l=W^g\mathbf{g}^T\mathcal{\Bar{X}}_l+b^g\text{, }\mathcal{X}_h=W^h\mathbf{h}^T\mathcal{\Bar{X}}_h+b^h\text{  ,}
\end{equation}
where $W^g,W^h\in\mathbb{R}^{C\times d}$ and $b^g,b^h\in\mathbb{R}^{d}$ are learnable parameters.
\subsection{Dual-Channel Spatio-Temporal Encoder}
The dual-channel spatio-temporal encoder, composed of the temporal attention, dilated causal convolution, and efficient spectral graph attention network (ESGAT).
\subsubsection{Disentangled Temporal Feature Extraction}
Different from previous works use a single method to model the intricate temporal patterns in the entangled traffic sequences, we use the dilated causal convolution and temporal attention to capture the short- and long-term temporal correlations in the high- and low-frequency components, respectively. The dilated causal convolution is a special 1D convolution, which slides over inputs by skipping values with a certain step, as illustrated in Figure \ref{model}. Mathematically, given a 1D sequence input $x\in\mathbb{R}^T$ and a filter $f\in\mathbb{R}^J$, the dilated causal convolution operation of $x$ with $f$ at time step $t$ is formulated as:
\begin{equation}
    x\star f(t)=\sum_{j=0}^Jf(j)x(t-c\times j)\text{  ,}
\end{equation}
where $c$ is the dilation factor. The dilated causal convolution for the high-frequency component is represented as:
\begin{equation}
    \mathcal{X}_h^{conv}=ReLU(\Theta\star \mathcal{X}_h+b)\text{  ,}
\end{equation}
where $\Theta$ and $b$ are learnable parameters, $ReLU(\cdot)$ is the rectified linear unit. Moreover, we utilize the masked self-attention on temporal dimension of the low-frequency component because the low-frequency component is stable and can represent the obvious long-term trends of traffic flow:
\begin{equation}
\begin{split}
    &\mathcal{X}_l^{tatt} = Concat(ta_1,...ta_n,...,ta_N)\\
    &where \quad ta_n = Att(X_l^n,X_l^n,X_l^n)\text{  .}
\end{split}
\end{equation}
\subsubsection{Efficient Global Spatial Feature Extraction}
For spatial correlation, we firstly consider adopting the vanilla GAT to dynamically calculate the weights between connected nodes. However, the spatial receptive field of the vanilla GAT is restricted in neighbors. Therefore, we further utilize the full GAT to dynamically capture the global spatial dependence by performing the self-attention on the spatial dimension of $\mathcal{X}^{conv}_h$ and $\mathcal{X}^{tatt}_l$. For simplicity, we remove the superscript and subscript in $\mathcal{X}^{conv}_h$ and $\mathcal{X}^{tatt}_l$ in this section. The full GAT with simplified input $\mathcal{X}$ can be formulated as:
\begin{equation}
\begin{split}
    &\mathcal{X}^{satt} = Concat(sa_1,...sa_t,...,sa_{T_1})\\
    &where \quad sa_t = Att(X_t,X_t,X_t)\text{  .}
\end{split}
\end{equation}
However, Eq. (9) is limited by the quadratic calculation complexity and lacks the structural information of the graph.
\paragraph{Query Sampling.} An intuitive way to reduce the complexity of Eq. (9) is to gain information from neighbors, which is equal to the vanilla GAT and loses the global information. To maintain these information, we propose a query sampling strategy to sample active nodes as sparse queries in the self-attention. The attention weights of unsampled nodes are copied from a sampled node which has the highest attention weight between them. This strategy is based on the fact that nodes in a region or community always have similar functions and weights under the hierarchical traffic system \cite{guo2021hierarchical}. Therefore, we utilize a GAT to pass message in the graph and a topk-pooling to sample active nodes which on behalf of regions or communities. The GAT can be formulated as:
\begin{equation}
\begin{split}
    &M_t = Concat(ma_1,...,ma_n,...,ma_N)\\
    &where\quad ma_n = Att(x^n_t,X^{\mathcal{N}_n}_{t},X^{\mathcal{N}_n}_{t})\text{  ,}
\end{split}
\end{equation}
where $\mathcal{N}_n$ and $M_t\in\mathbb{R}^{N\times d}$ denote the index of neighbors of node $n$ and the scores of nodes at time step $t$. Then we utilize the topk-pooling to sample $\lceil logN\rceil$ active nodes that receive max flow from other nodes in the GAT. Specifically, to evaluate how much flow from other nodes can be retained, we employ a trainable projection vector $P\in\mathbb{R}^{d\times 1}$ to project the score matrix to 1D and sample nodes according to values:
\begin{equation}
    idx_t = rank(\frac{M_tP}{\Vert P\Vert}, \lceil logN\rceil)\text{  ,}
\end{equation}
where $rank(\cdot)$ returns the index of the top $\lceil logN\rceil$ largest values, and $idx_t\in\mathbb{R}^{\lceil logN\rceil}$ indicates the index of sampled queries at time step $t$. Finally, Eq. (9) can be rewritten as an efficient version with query sampling:
\begin{equation}
\begin{split}
    &\mathcal{X}^{egat} = Concat(sa_1,...sa_t,...,sa_{T_1})\\
    &where \quad sa_t = Att(X_t^{idx_t},X_t,X_t)\text{  .}
\end{split}
\end{equation}
\paragraph{Graph Positional Encoding.} To effectively guide message passing in the full GAT, we propose a novel graph positional encoding. In the vanilla self-attention, sine and cosine functions are utilized as the positional encoding for sequences, which is a essential part of the vanilla self-attention. However, sinusoids cannot be clearly defined in graphs, since there is no clear notion of position along an axis. Previous works utilize graph embedding algorithms to generate vectors as the graph positional encoding, but they are limited by the local context with extra parameters. Furthermore, \cite{dwivedi2020generalization} uses graph Laplacian eigenvectors as the graph positional encoding because the eigenvectors of the graph Laplacian are the natural equivalent of sine functions, which can reveal the structural information in the graph. However, the influence of the eigenvectors on the signal of one node is not localized in its neighborhood \cite{DBLP:conf/iclr/XuSCQC19}. Inspired by the graph wavelet transform, our proposed graph positional encoding is the wavelet bases of the graph. The graph wavelet corresponds to graph Laplacian eigenvectors diffused away from a centered node with a scaling matrix on the graph and can reflect the localization property compared with eigenvectors. The wavelet-based graph positional encoding can be formulated as:
\begin{equation}
    \rho^{spa}=\Phi^{spa}G_{s}^{spa^{\frac{1}{2}}}\text{, }\rho^{tem}=\Phi^{tem}G_{s}^{tem^{\frac{1}{2}}}\text{  ,}
\end{equation}
where $\rho^{spa},\rho^{tem}\in\mathbb{R}^{N\times d}$ are the graph positional encodings of the spatial graph and the temporal graph in our model. The dot-product of $\rho$ and its transpose $\rho^T$ is equivalent to wavelets of graphs, which can show not only the structural information but also the localization property of graphs. We further set the scale $s$ in Eq. (13) as the learnable parameter to avoid misleading inductive bias. Finally, our efficient spectral graph attention can be formulated as follows:
\begin{equation}
\begin{split}
    &\mathcal{X}^{esgat} = Concat(sa_1,...sa_t,...,sa_{T_1})\\
    &where \quad sa_t = Att(\tilde{X}_t^{idx_t},\tilde{X}_t,\tilde{X}_t)\\
    &and\quad \tilde{\mathcal{X}}=\mathcal{X}+\rho^{spa}+\rho^{tem}\text{  .}
\end{split}
\end{equation}

%

\subsection{Frequency-Specific Decoder}
To transform the representations encoded by the dual-channel encoder into the future for the multi-step traffic flow forecasting, we use the predictor (\emph{i.e.}, a fully-connected layer) on the temporal dimension of $\mathcal{X}^{esgat}_l,\mathcal{X}^{esgat}_h\in\mathbb{R}^{T_1\times N\times d}$ to derive the future representations $\mathcal{\hat{Y}}^{f}_l,\mathcal{\hat{Y}}^{f}_h\in\mathbb{R}^{T_2\times N\times d}$ of the low- and high-frequency components. Then we utilize the fusion attention and multi-supervision to merge the information of the low- and high-frequency components and gain knowledge by supervising the low-frequency component.
\subsubsection{Disentangled Temporal Feature Fusion}
The aim of our paper is not to forecast the low- and high-frequency components but the traffic flow in the future. Therefore, as shown in Figure \ref{model}, we further propose a fusion attention to not only merge the representations of the low- and high-frequency components $\mathcal{\hat{Y}}^{f}_l,\mathcal{\hat{Y}}^{f}_h$ into the traffic flow $\mathcal{\hat{Y}}^{f}\in\mathbb{R}^{T_2\times N\times d}$ but also capture the intra-dependencies in the future. Specifically, the fusion attention regards the low-frequency component as queries in two temporal attentions to extract useful long- and short-term information from the low- and high-frequency components. Fusion attention can be formulated as :

\begin{small}
\begin{equation}
\begin{split}
    &\mathcal{\hat{Y}}^f = Concat(fa_1,...,fa_n,...,fa_N)\\
    &where\quad fa_n = Att(\hat{Y}^{f^n}_l,\hat{Y}^{f^n}_l,\hat{Y}^{f^n}_l)\\
    &\quad\quad\quad\quad\quad\quad +Att(\hat{Y}^{f^n}_l,\hat{Y}^{f^n}_h,\hat{Y}^{f^n}_h)\text{  .}
\end{split}
\end{equation}
\end{small}
\subsubsection{Multi-Supervision}
We use a fully-connected layer to transform the future representations of traffic flow $\mathcal{\hat{Y}}^f$ into the expected prediction $\mathcal{\hat{Y}}$ with $L1$ loss during training. Besides, we supervise the low-frequency component in our model same as the traffic flow. By gaining knowledge from the much stable low-frequency component, our model can effectively enhance its capability of learning the long-term trends of traffic flow, thus yielding better performance. Therefore, STWave is optimized by minimizing the following loss function:
\begin{equation}
    \mathcal{L}=\sum_{t=T_1+1}^{T_1+T_2}\sum_{n=1}^N\vert x^n_t-\hat{y}^n_t\vert+\vert x^n_{l_t}-\hat{y}^n_{l_t}\vert\text{  .}
\end{equation}
\subsection*{Complexity Analysis}
The complexity of the dual-channel encoder is $O(L(TNJ+NT^2+TNlogN))$, where dilated causal convolution, temporal attention, ESGAT cost $O(TNJ)$, $O(NT^2)$, $O(TNlogN)$ complexity, and $L$ denotes the number of stacked layers. The complexity of the disentangling layer and decoder is $O(NT)$ and $O(NT^2)$. Therefore, STWave achieves comparable time complexity as compared to other GCN-based frameworks.
\begin{table}[t]
    \centering
    \renewcommand\arraystretch{1.0}
    \resizebox{1.0\linewidth}{!}{\begin{tabular}{lllllll}
    \toprule[1.5pt]
         Dataset & \#Nodes & \#Edges & \#Samples & Sample Rate & \#MissingRatio & Time range\\
    \midrule[1pt]
         PeMSD3 & 358 & 547 & 26208 & 5 mins & 0.672\% & 9/1/2018-11/30/2018\\
         PeMSD4 & 307 & 340 & 16992 & 5 mins & 3.182\% & 1/1/2018-2/28/2018\\
         PeMSD7 & 883 & 866 & 28224 & 5 mins & 0.452\% & 5/1/2017-8/31/2017\\
         PeMSD8 & 170 & 295 & 17856 & 5 mins & 0.696\% & 7/1/2016-8/3/2016\\
    \bottomrule[1.5pt]
    \end{tabular}}
    \caption{Dataset statistics}
    \label{dataset}
\end{table}
\begin{table*}[t]
    \aboverulesep=0ex
    \belowrulesep=0ex
    \centering
    \resizebox{1.0\linewidth}{!}{\begin{tabular}{l|cccccccccccc}
    \toprule[1pt]
        \multirow{2}{*}{\large Methods} & \multicolumn{3}{c}{PeMSD3} & \multicolumn{3}{c}{PeMSD4} & \multicolumn{3}{c}{PeMSD7} & \multicolumn{3}{c}{PeMSD8}\\
        \cmidrule(lr){2-4} \cmidrule(lr){5-7} \cmidrule(lr){8-10} \cmidrule(lr){11-13}
         & MAE & RMSE & MAPE & MAE & RMSE & MAPE & MAE & RMSE & MAPE & MAE & RMSE & MAPE\\
    \midrule[0.5pt] 
        HA & 31.58 & 52.39 & 33.78\% & 38.03 & 59.24 & 27.88\% & 45.12 & 65.64 & 24.51\% & 34.86 & 59.24 & 27.88\%\\
        ARIMA & 35.41 & 47.59 & 33.78\% & 33.73 & 48.80 & 24.18\% & 38.17 & 59.27 & 19.46\% & 31.09 & 44.32 & 22.73\%\\
        VAR & 23.65 & 38.26 & 24.51\% & 24.54 & 38.61 & 17.24\% & 50.22 & 75.63 & 32.22\% & 19.19 & 29.81 & 13.10\%\\
        SVR & 21.97 & 35.29 & 21.51\% & 28.70 & 44.56 & 19.20\% & 32.49 & 50.22 & 14.26\% & 23.25 & 36.16 & 14.64\%\\
        LSTM & 21.33 & 35.11 & 23.33\% & 26.77 & 40.65 & 18.23\% & 29.98 & 45.94 & 13.20\% & 23.09 & 35.17 & 14.99\%\\
        TCN & 19.32 & 33.55 & 19.93\% & 23.22 & 37.26 & 15.59\% & 32.72 & 42.23 & 14.26\% & 22.72 & 35.79 & 14.03\%\\
        STGCN & 17.55 & 30.42 & 17.34\% & 21.16 & 34.89 & 13.83\% & 25.33 & 39.34 & 11.21\% & 17.50 & 27.09 & 11.29\%\\
        DCRNN & 17.99 & 30.31 & 18.34\% & 21.22 & 33.44 & 14.17\% & 25.22 & 38.61 & 11.82\% & 16.82 & 26.36 & 10.92\%\\
        Graph WaveNet & 19.12 & 32.77 & 18.89\% & 24.89 & 39.66 & 17.29\% & 26.39 & 41.50 & 11.97\% & 18.28 & 30.05 & 12.15\%\\
        ASTGCN(r) & 17.34 & 29.56 & 17.21\% & 22.93 & 35.22 & 16.56\% & 24.01 & 37.87 & 10.73\% & 18.25 & 28.06 & 11.64\%\\
        LSGCN & 17.94 & 29.85 & 16.98\% & 21.53 & 33.86 & 13.18\% & 27.31 & 41.46 & 11.98\% & 17.73 & 26.76 & 11.20\%\\
        STSGCN & 17.48 & 29.21 & 16.78\% & 21.19 & 33.65 & 13.90\% & 24.26 & 39.03 & 10.21\% & 17.13 & 26.80 & 10.96\%\\
        AGCRN & \underline{15.98} & \underline{28.25} & \underline{15.23}\% & \underline{19.83} & \underline{32.26} & \underline{12.97}\% & \underline{22.37} & \underline{36.55} & \underline{9.12}\% & \underline{15.95} & \underline{25.22} & \underline{10.09}\%\\
        STFGNN & 16.77 & 28.34 & 16.30\% & 20.48 & 32.51 & 16.77\% & 23.46 & 36.60 & 9.21\% & 16.94 & 26.25 & 10.60\%\\
        STGODE & 16.50 & 27.84 & 16.69\% & 20.84 & 32.82 & 13.77\% & 22.59 & 37.54 & 10.14\% & 16.81 & 25.97 & 10.62\%\\
    \midrule[0.5pt] 
        STWave & \textbf{14.93} & \textbf{26.50} & \textbf{15.05}\% & \textbf{18.50} & \textbf{30.39} & \textbf{12.43}\% & \textbf{19.94} & \textbf{33.88} & \textbf{8.38}\% & \textbf{13.42} & \textbf{23.40} & \textbf{8.90}\%\\
    \bottomrule[1pt]
    \end{tabular}}
    \caption{Comparison of STWave and baselines on four traffic datasets. \textbf{Bold}: Best, \underline{underline}: Second best.}
    \label{mainres}
\end{table*}
\section{Experiments}
We investigate the effectiveness of STWave with the goal of answering five research questions. RQ1: does our proposed STWave outperform the baselines? RQ2: how do different components of STWave (\emph{e.g.}, ESGAT) affect its performance? RQ3: how do hyper-parameters affect STWave? RQ4: does our proposed STWave more efficient than baselines? RQ5: how does wavelet affect STWave?
\subsection{Experimental Setup}
\paragraph{Datasets and Metrics.} We evaluate our model on four real-world datasets collected from the California Transportation Agencies (CalTrans) Performance Measurement System (PeMS). Descriptive statistics for those datasets are presented in Table \ref{dataset}. Following \cite{guo2019attention}, we use the observations flow from the previous $12$ time steps to predict the next $12$ steps and split these four datasets into a training set (60\%), validation set (20\%), and test set (20\%) in chronological order. Three standard metrics are adopted to evaluate the performance of all methods, namely, Mean Absolute Errors (MAE), Mean Absolute Percentage Errors (MAPE), and Root Mean Squared Errors (RMSE).
\paragraph{Baselines.} We compare our proposed STWave with the following baseline models in conjunction with the models we introduced in the related work$\footnote{Related work is in Appendix A due to space limitation.}$ section — in total, we use $15$ baseline models: 1) HA \cite{hamilton2020time}: uses the average value of the last $12$ times slices to predict the next value; 2) ARIMA \cite{williams2003modeling}: a statistical model of time series analysis; 3) VAR \cite{lu2016integrating}: a time series model that can capture spatial correlations among all nodes; 4) SVR \cite{wu2004travel}: utilizes a linear support vector machine to perform regression; 5) STSGCN \cite{song2020spatial}: uses a spatial-temporal synchronous mechanism to capture the localized spatial-temporal correlations.
\paragraph{Parameter Settings.} We implement STWave using PyTorch and train model for $200$ epochs using the Adam optimizer, with a batch size of $64$ on all datasets. We present the default hyper-parameter settings of STWave as follows: the number of head $e$ and dimension $d_e$ of each head in attention is set as $8$ and $16$ (all attention modules in STWave is implemented by multi-head mechanism). Besides, the number of layers $L$ in spatio-temporal encoder is set as $2$. We integrate high-frequency temporal dependencies by stack $1$ dilated causal convolution layer with kernel size $J=2$. The learning rate is initialized as $0.001$ with a $0.1$ decay rate.
\subsection{Performance Comparison (RQ1)}
The results of all methods across four datasets for forecasting traffic flow (measured by MAE, RMSE, and MAPE) are reported in Table \ref{mainres}, and results for each time step are shown in Appendix B. From Table \ref{mainres}, we have the following observations. HA provides a lower bound of model performance. The results of traditional and machine learning methods (ARIMA, VAR, and SVR) are much worse than the deep models due to the non-linear dependencies and lack of hand-craft features. In terms of all tasks, the non-graph-based methods (LSTM and TCN) generally perform worse than graph-based baselines (\emph{e.g.}, STGCN, DCRNN, and Graph WaveNet), demonstrating the efficacy of graph on modeling the spatial dependence. As for graph-based methods, ASTGCN and LSGCN perform better than Graph WaveNet, which indicates the effectiveness of the attention mechanism in modeling dynamic interactions between roads. STFGNN and STGODE perform better than prior graph-based methods, as they carefully design the temporal graph and graph ordinary differential equation to expand the receptive field of GCN, which, however, are inferior than AGCRN, due to they fail to capture the global spatial dependence. Overall, our STWave achieves the best performance on all tasks. This is mainly because: 1) STWave extracts the low- and high-frequency components from traffic sequences and utilizes the dual-channel encoder to model each component individually; 2) STWave adopts the fusion attention and multi-supervision to fully merge and exploit the high- and low-frequency information of traffic series; and 3) STWave develops the powerful wavelet-based graph positional encoding in full GAT to effectively capture global dependence by injecting structural information.
\subsection{Ablation Study (RQ2)}
\begin{figure}[t]
    \centering
        \begin{subfigure}{0.33\linewidth}
        \includegraphics[width=\linewidth]{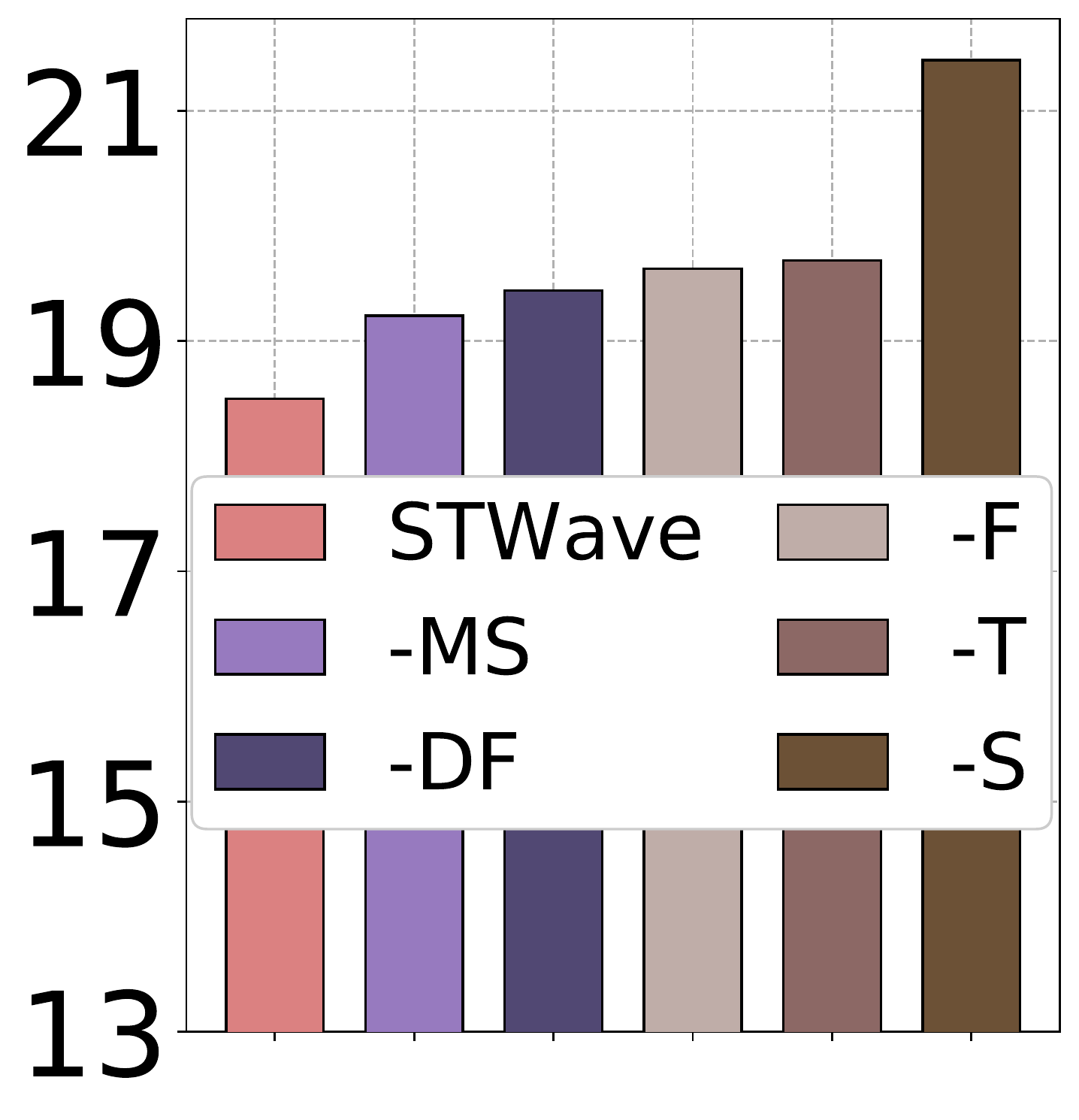}
        \captionsetup{font=small}
        \caption{PeMSD4 MAE}
      \end{subfigure}%
      \hfill
      \begin{subfigure}{0.33\linewidth}
        \includegraphics[width=\linewidth]{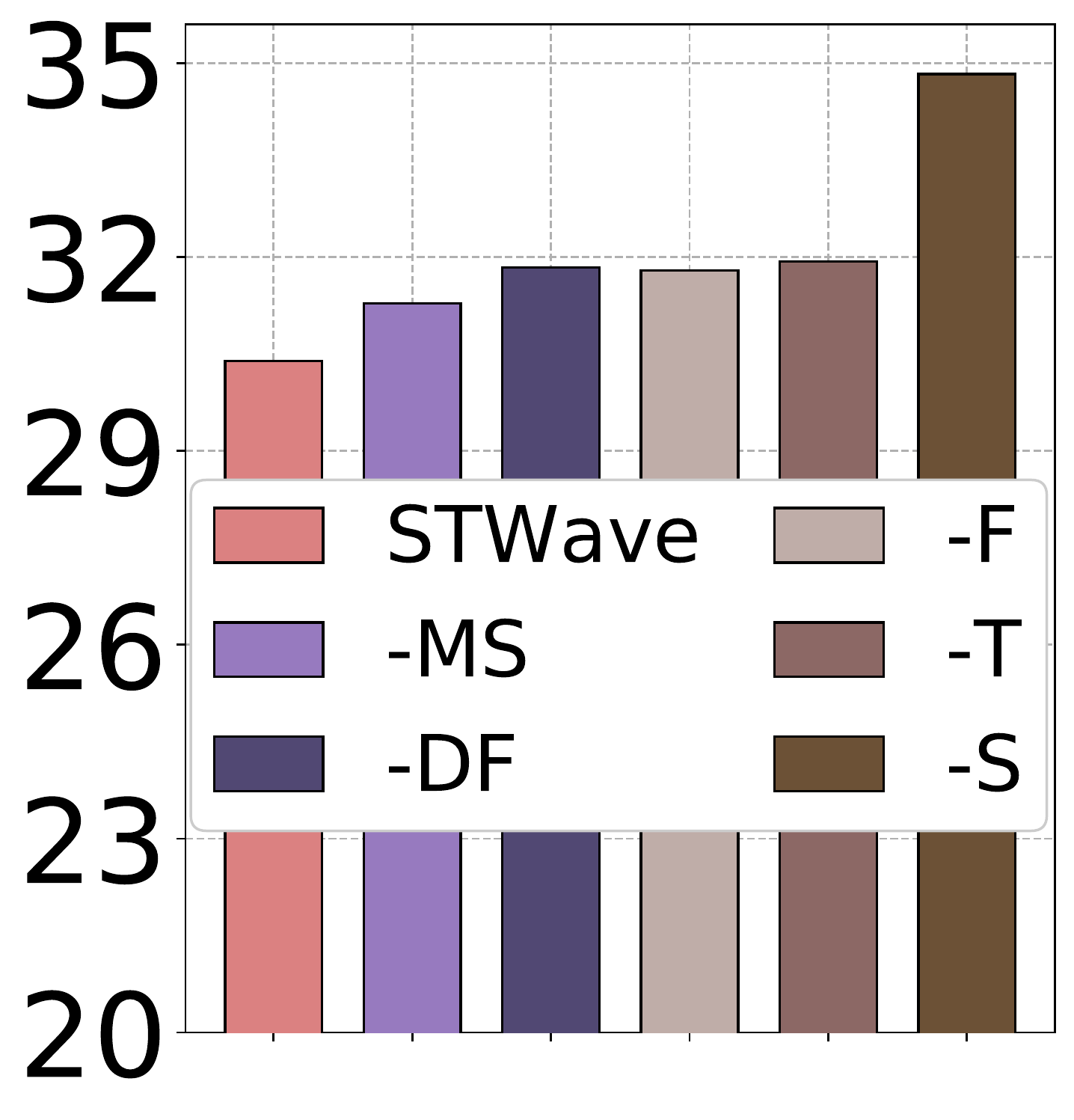}
        \captionsetup{font=small}
        \caption{PeMSD4 RMSE}
      \end{subfigure}
      \hfill
      \begin{subfigure}{0.33\linewidth}
        \includegraphics[width=\linewidth]{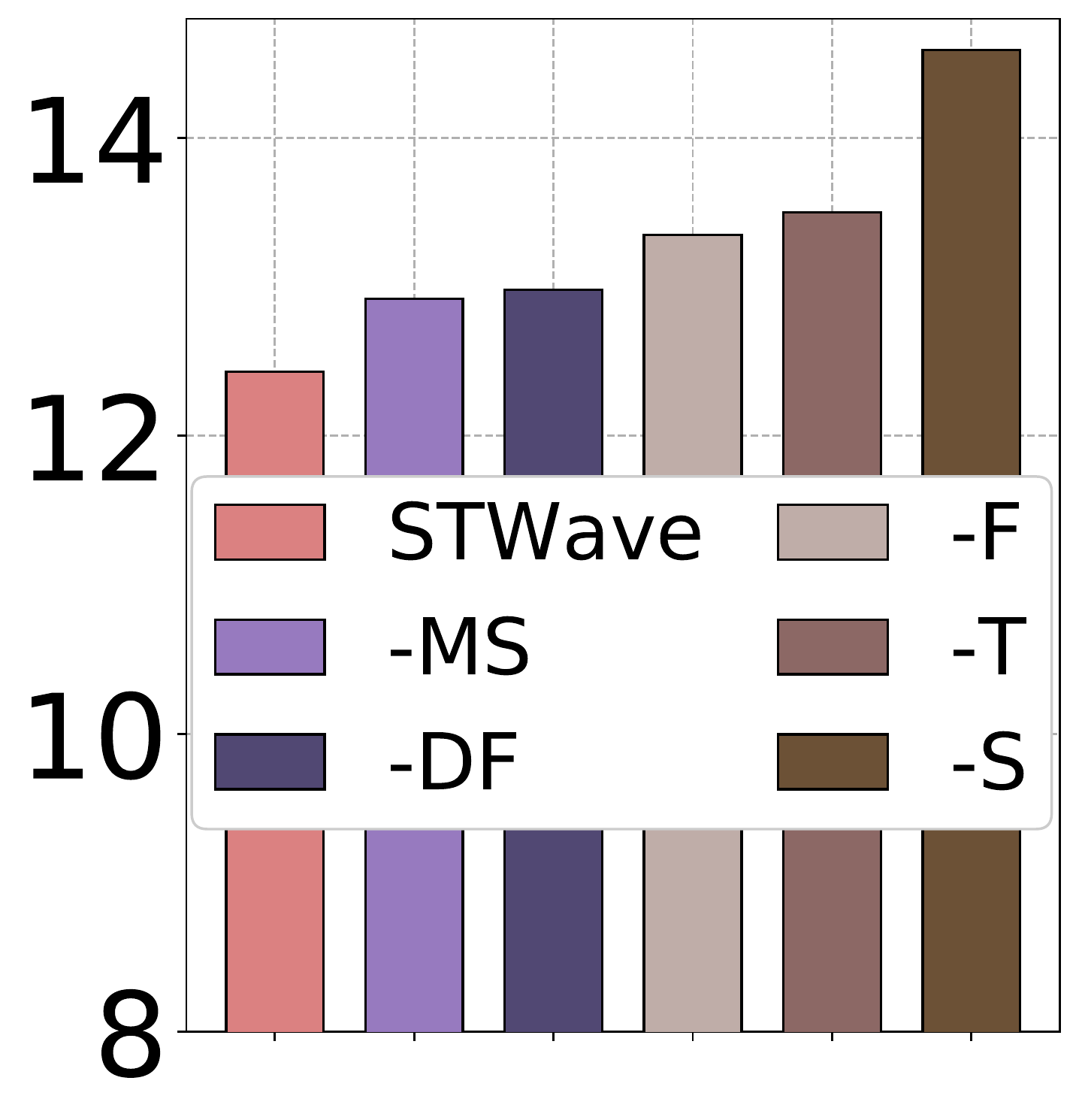}
        \captionsetup{font=small}
        \caption{PeMSD4 MAPE}
      \end{subfigure}
      
      \begin{subfigure}{0.33\linewidth}
        \includegraphics[width=\linewidth]{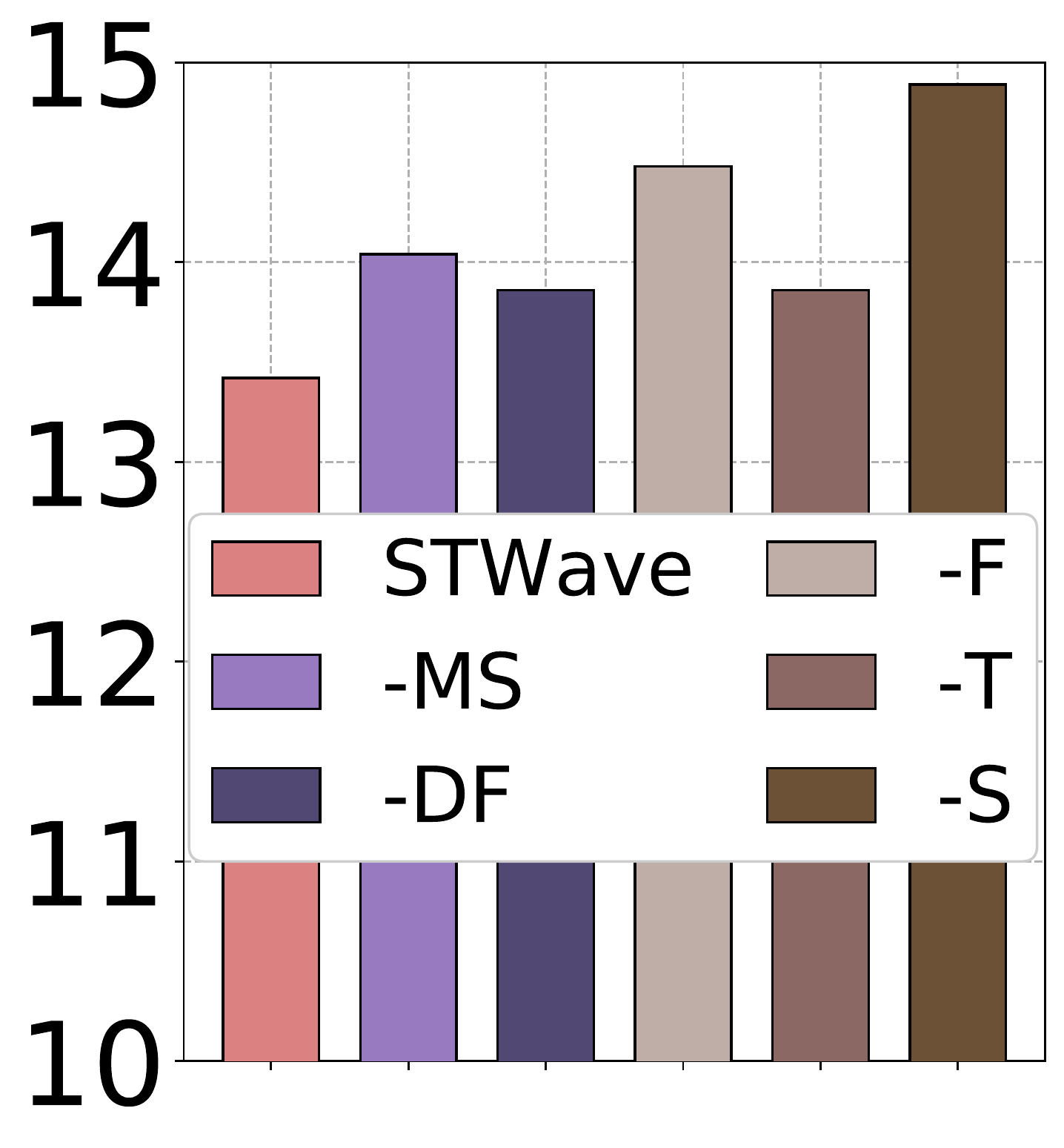}
        \captionsetup{font=small}
        \caption{PeMSD8 MAE}
      \end{subfigure}%
      \hfill
      \begin{subfigure}{0.33\linewidth}
        \includegraphics[width=\linewidth]{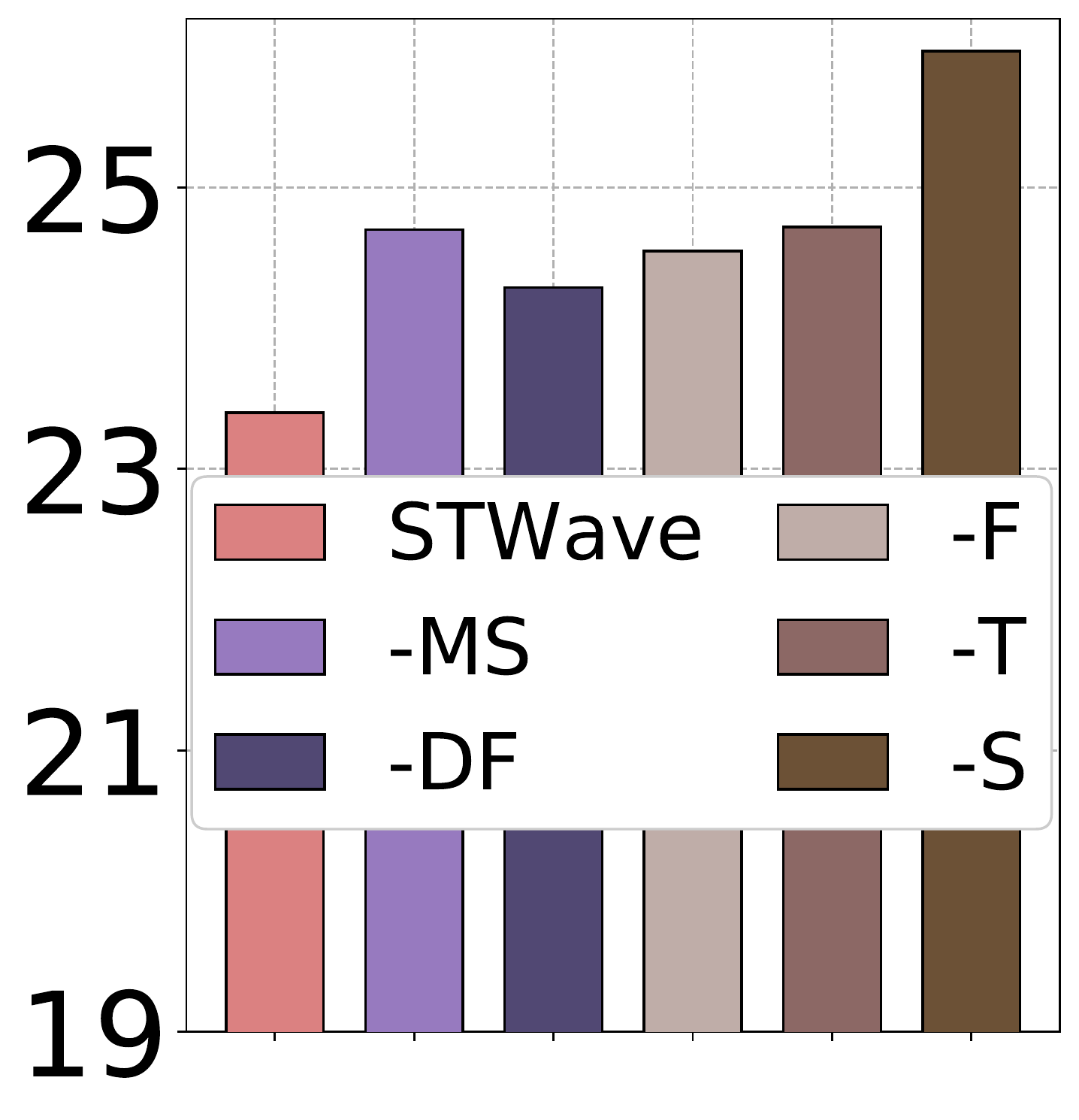}
        \captionsetup{font=small}
        \caption{PeMSD8 RMSE}
      \end{subfigure}
      \hfill
      \begin{subfigure}{0.33\linewidth}
        \includegraphics[width=\linewidth]{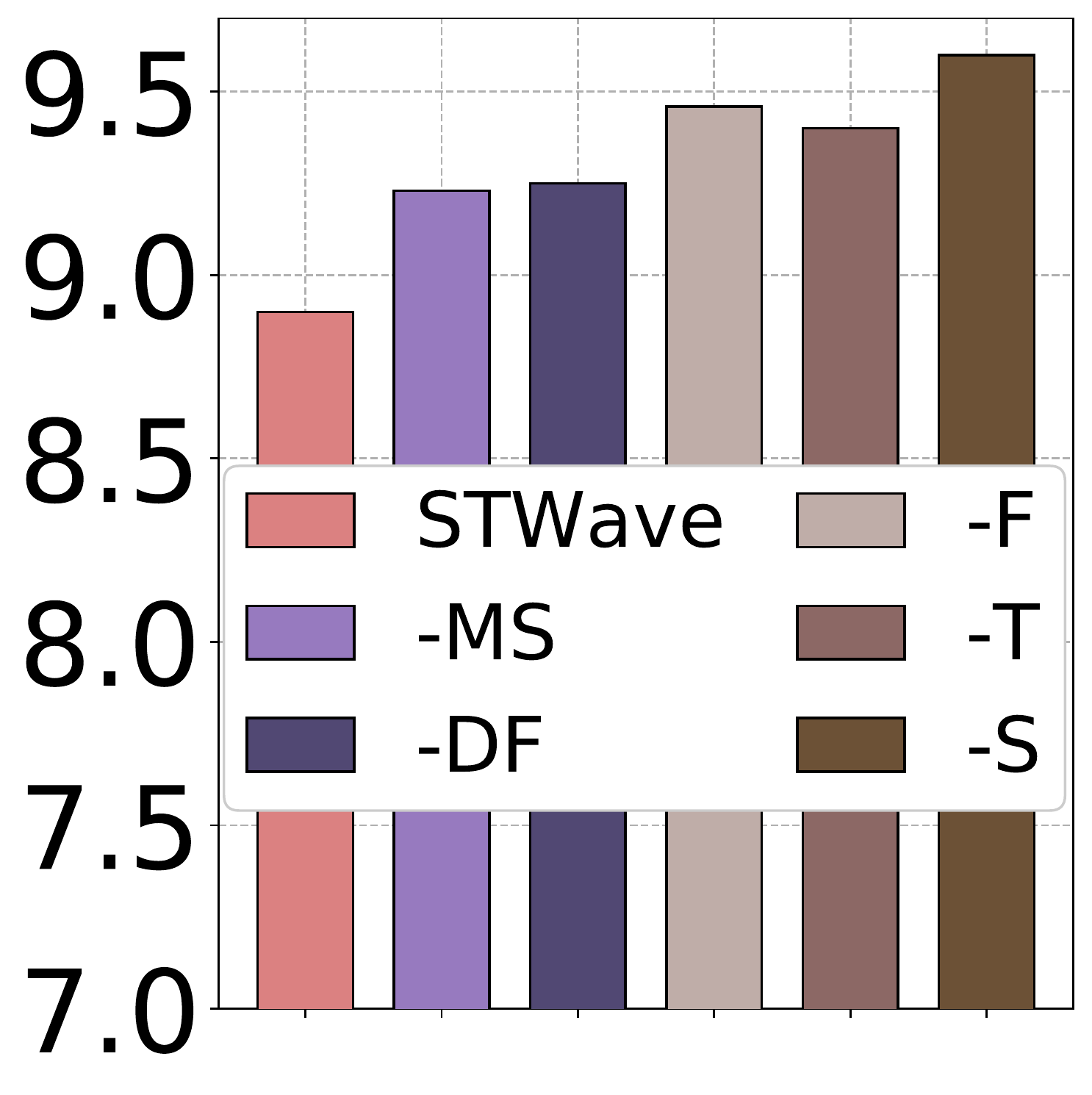}
        \captionsetup{font=small}
        \caption{PeMSD8 MAPE}
      \end{subfigure}
      \caption{Performance comparison for variants of STWave.}
      \label{abl}
\end{figure}

To investigate the effectiveness of different components of STWave, we compare it with five different variants: 1) "-MS": STWave without the multi-supervision; 2) "-DF": STWave without the disentangling flow layer; 3) "-F": STWave replaces the fusion attention with the addition operation; 4) "-T": STWave without the dilated causal convolution and temporal attention; 5) "-S": STWave without the ESGAT. Figure \ref{abl} shows the comparison results. It is obvious that the full version of STWave achieves the best performance compared to variants. Generally, the performance decrease of "-S" far exceeds that of "-T", implying that the spatial dimension plays a more important role than the temporal dimension in our model. Both "-F", "-DF", and "-MS" underperform STWave, indicating the advantages of modeling intricate patterns individually. In summary, our proposed STWave benefits from the five delicately-designed components.
\begin{figure}[t]
    \centering
        \begin{subfigure}{0.33\linewidth}
        \includegraphics[width=\linewidth]{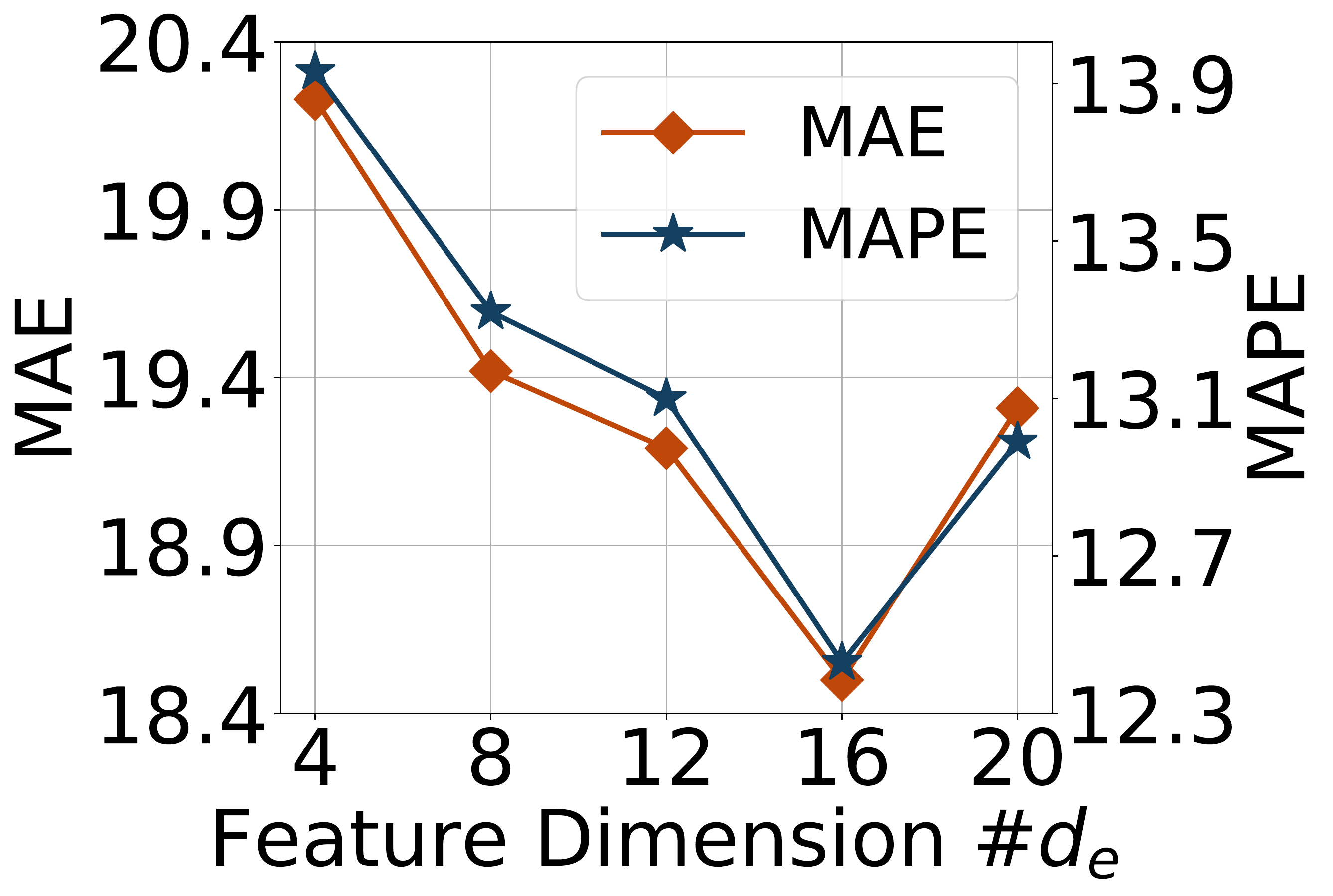}
        \captionsetup{font=small}
      \end{subfigure}%
      \hfill
      \begin{subfigure}{0.33\linewidth}
        \includegraphics[width=\linewidth]{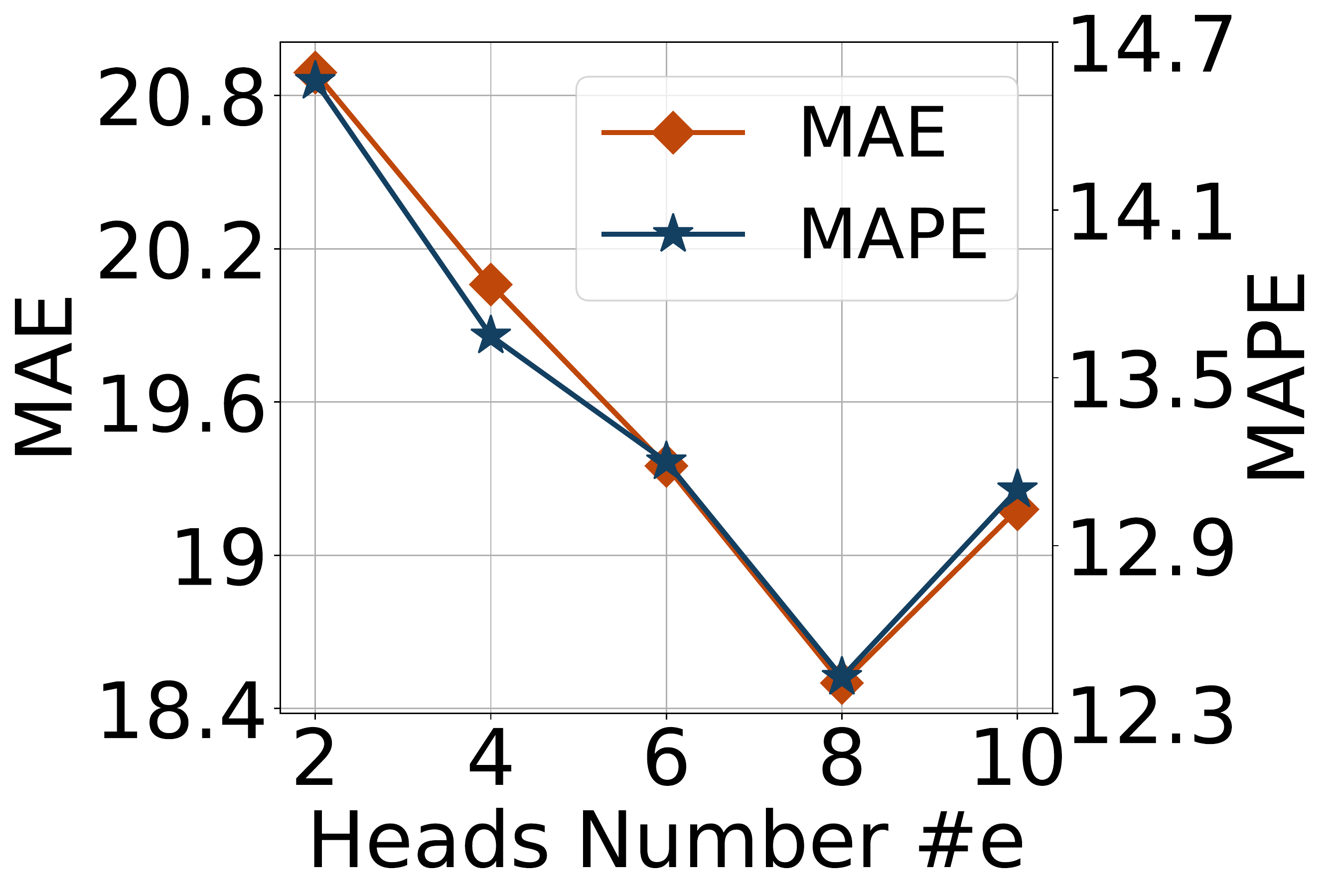}
        \captionsetup{font=small}
      \end{subfigure}
      \hfill
      \begin{subfigure}{0.33\linewidth}
        \includegraphics[width=\linewidth]{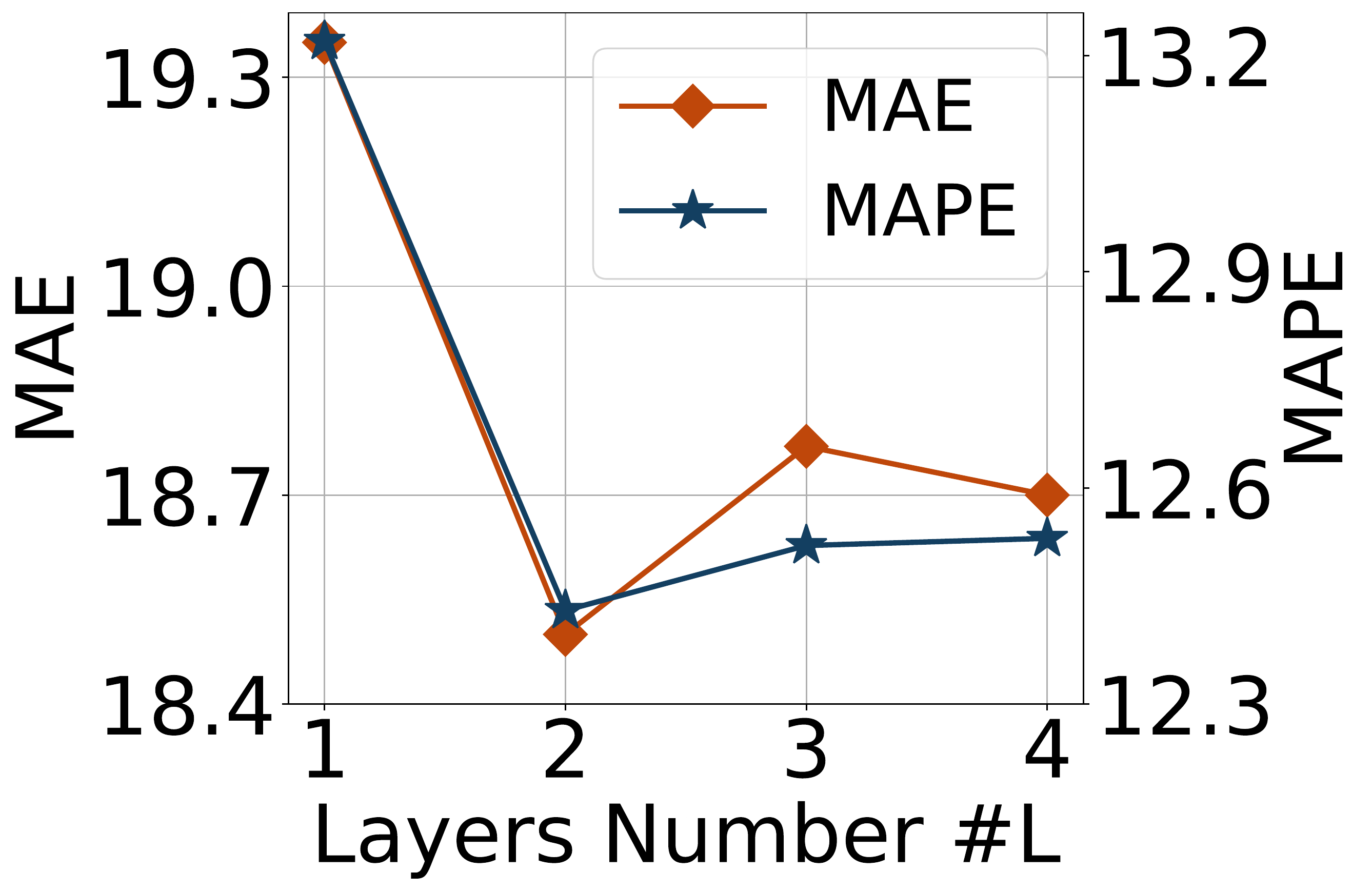}
        \captionsetup{font=small}
      \end{subfigure}
      
      \begin{subfigure}{0.33\linewidth}
        \includegraphics[width=\linewidth]{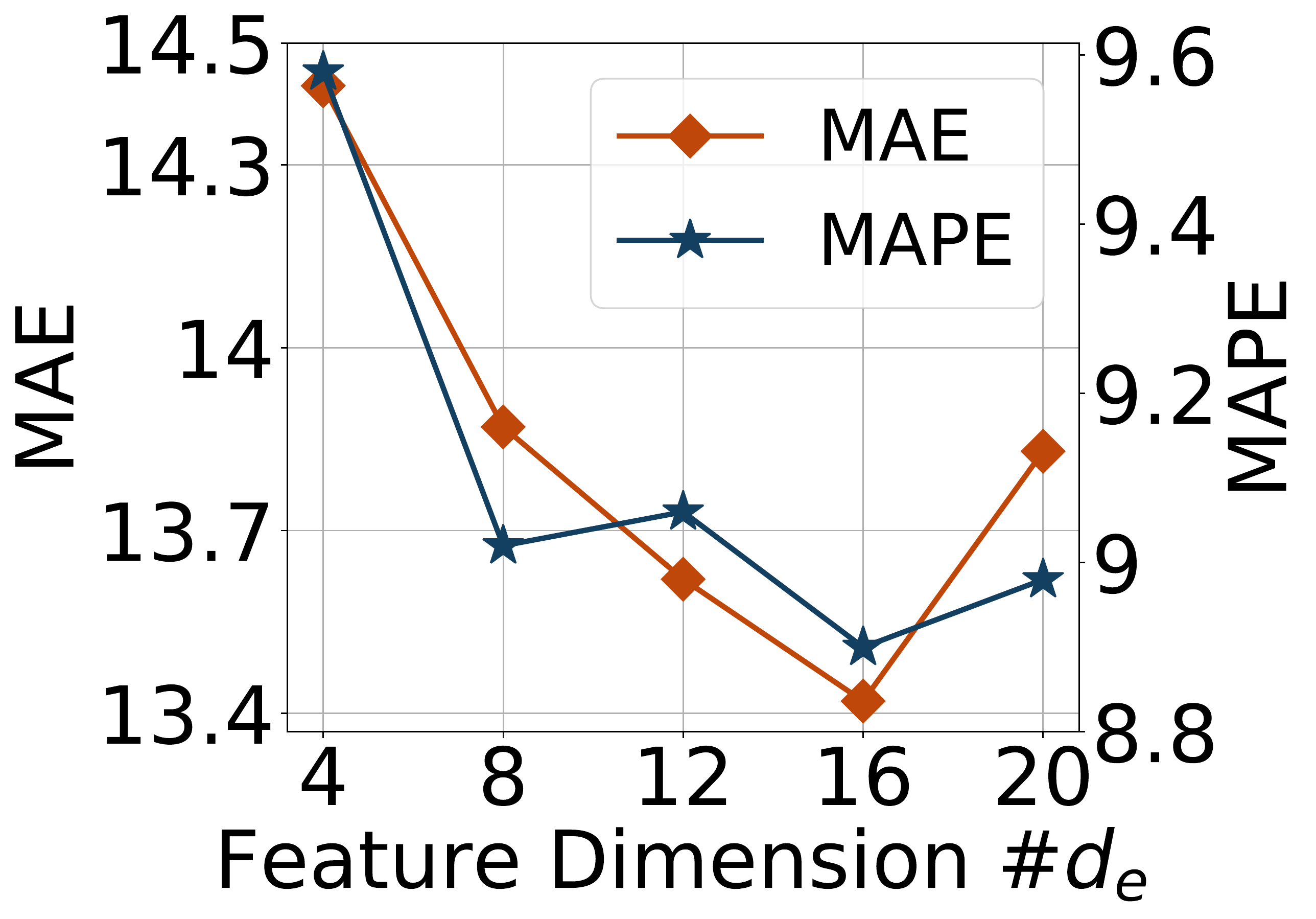}
        \captionsetup{font=small}
      \end{subfigure}%
      \hfill
      \begin{subfigure}{0.33\linewidth}
        \includegraphics[width=\linewidth]{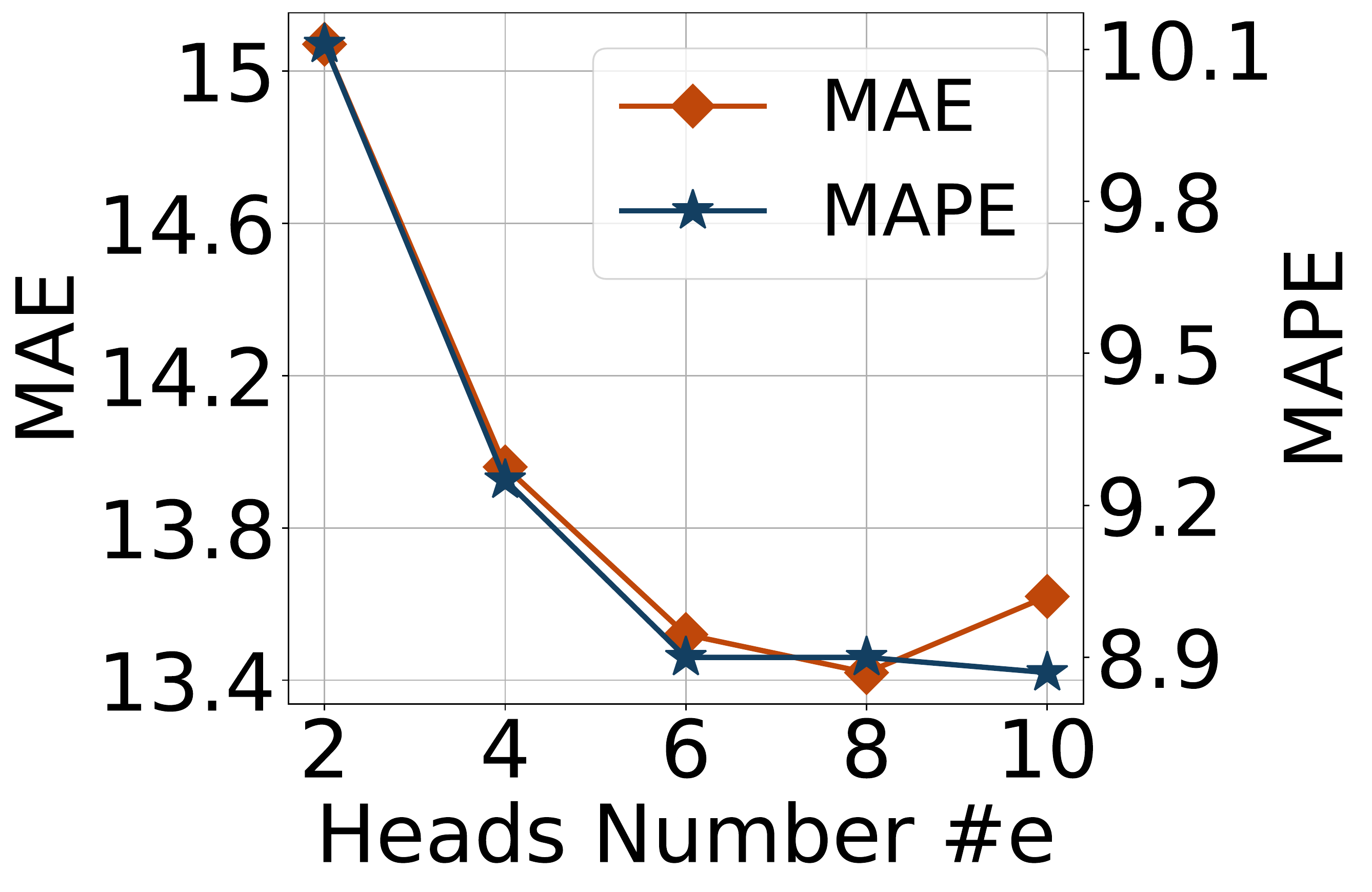}
        \captionsetup{font=small}
      \end{subfigure}
      \hfill
      \begin{subfigure}{0.33\linewidth}
        \includegraphics[width=\linewidth]{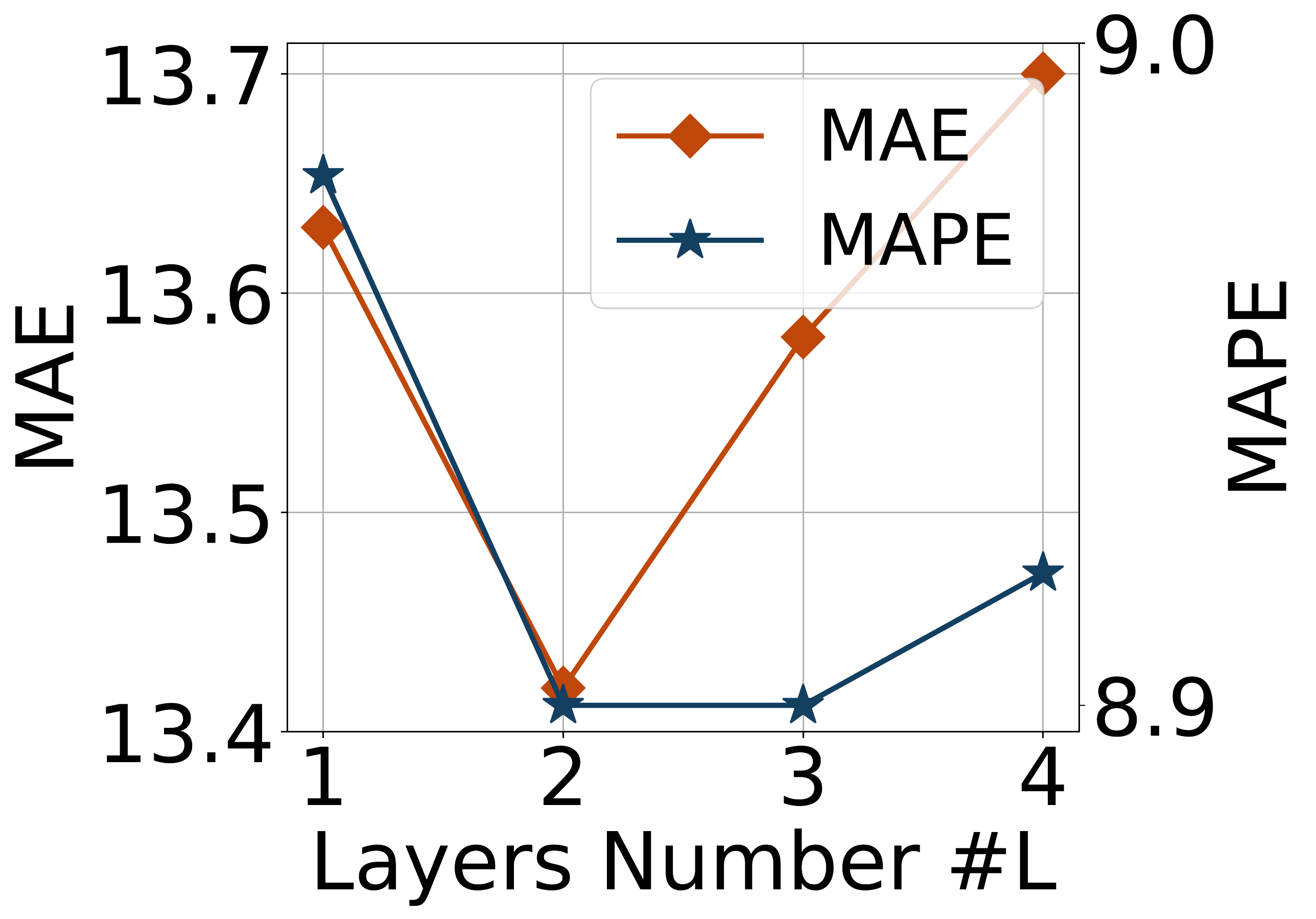}
        \captionsetup{font=small}
      \end{subfigure}
      \caption{Hyper-parameter study on PeMSD4 and D8 datasets.}
      \label{param}
\end{figure}
\subsection{Parameter Sensitivity Analysis (RQ3)}
Figure \ref{param} depicts the results of parameter sensitivity analysis on traffic forecasting. We search the number of heads and dimensions of each head in attention from a search space of [$2,4,6,8,10$] and [$4,8,12,16,20$]. For the head number, STWave with $8$ heads in attention outperforms the others. For the dimension of each head, the best performance is achieved with $16$. Clearly, increasing the model size is capable of endowing our predictive model with better representation ability. Increasing more heads and dimensions may involve noise in refining learned representations. Furthermore, the performance of STWave climbs up as the number of layers increases and becomes stable with the number around $2$.
\begin{figure}[t]
    \centering
        \begin{subfigure}{0.49\linewidth}
        \includegraphics[width=\linewidth]{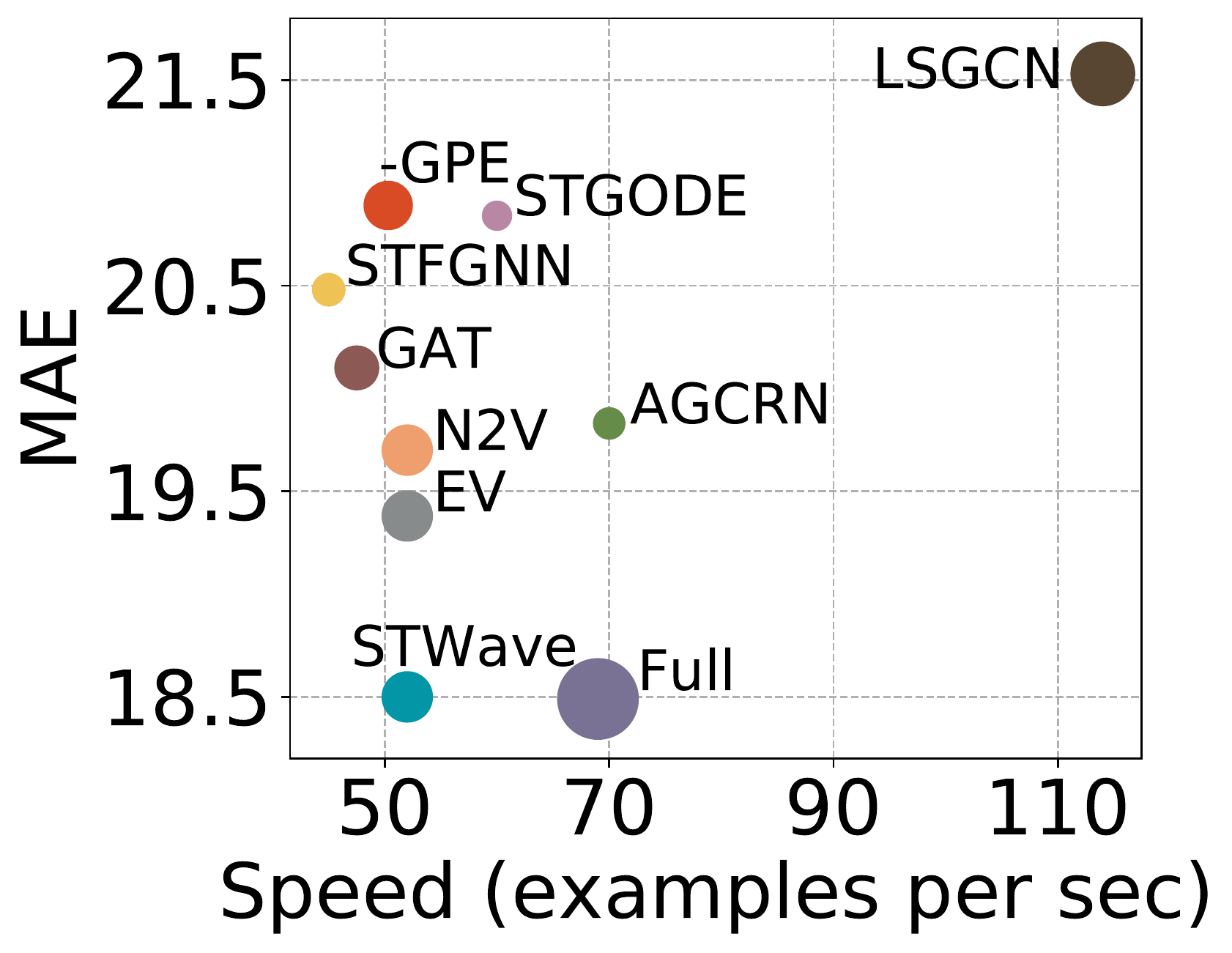}
        \captionsetup{font=small}
        \caption{Performance ($y$ axis), speed ($x$ axis), and memory footprint (size of the circles) of methods.}
        \label{performance}
      \end{subfigure}%
      \hfill
      \begin{subfigure}{0.49\linewidth}
        \includegraphics[width=\linewidth]{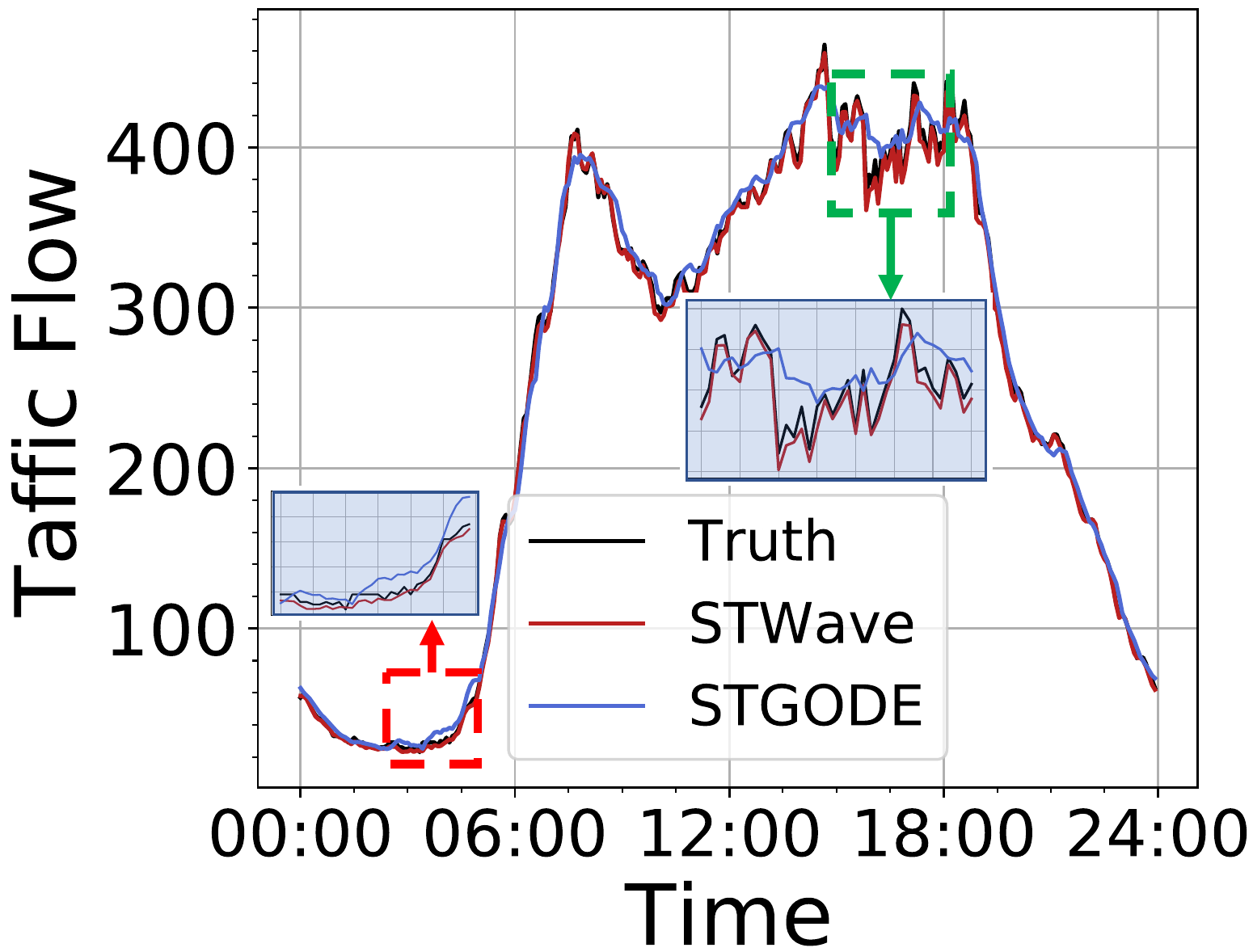}
        \captionsetup{font=small}
        \caption{Traffic forecasting visualization. Traffic flow of node 261 on 2/27/2018.}
        \label{vis}
      \end{subfigure}
      
      \begin{subfigure}{0.8\linewidth}
        \includegraphics[width=\linewidth]{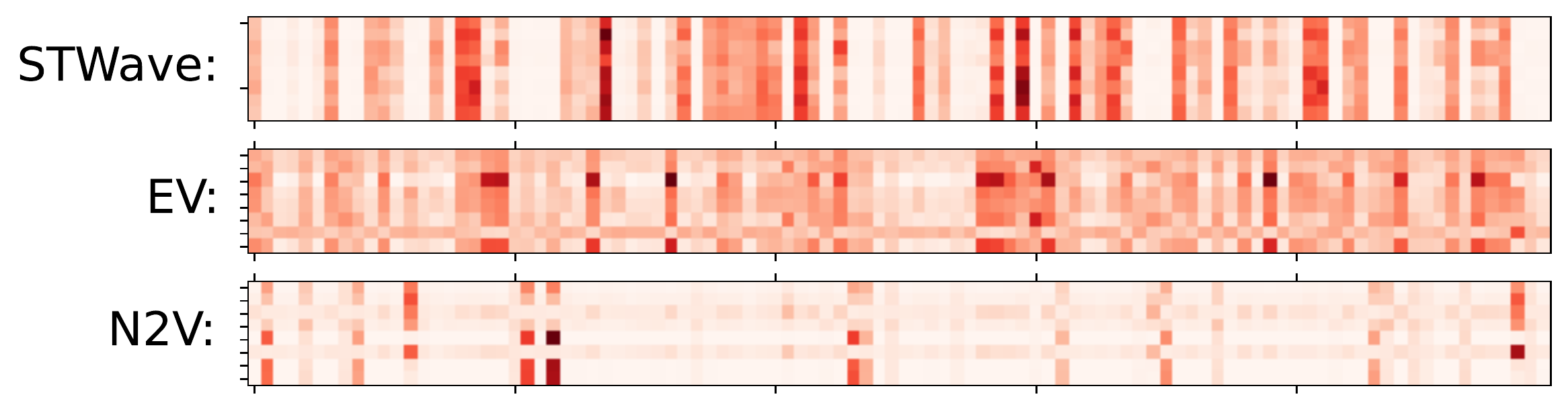}
        \captionsetup{font=small}
        \caption{Visualization for the learned weights on 2/27/2018.}
        \label{visadj}
      \end{subfigure}
    \caption{Wavelet study on the PeMSD4 dataset.}
\end{figure}
\subsection{Computation Cost (RQ4)}
To show the performance on computation of our model, we compare STWave with the attention-based LSGCN, state-of-the-art baselines, and two different variants of STWave: 1) "GAT": STWave replaces ESGAT with the vanilla GAT; 2) "Full": STWave without the query sample in ESGAT. Figure \ref{performance} shows the trade-off between qualitative performance, model speed, and memory footprint. While "Full" performs well, its speed and memory footprint is larger than STWave. On the other hand, STFGNN and "GAT" is fast at the cost of lower quantitative performance. Prior attention-based LSGCN are worst in all aspects. Among these models, STWave seems to be able to make a better trade-off in terms of speed and performance, while having reasonable memory usage.
\subsection{Wavelet Study (RQ5)}
To investigate the effectiveness of graph wavelet-based graph positional encoding, we design three variants of STWave: 1) "-GPE": without the graph positional encoding; 2) "EV": sets the eigenvectors of the graph Laplacian as the graph positional encoding; 3) "N2V": sets the matrix derived by node2vec \cite{grover2016node2vec} as the graph positional encoding. As shown in Figure \ref{performance}, "-GPE" has worst performance due to it without the inductive bias of structural information. The performance of "EV" and "N2V" are worse than STWave due to they fail to consider both the structural information and localization property. The attention weights of them are visualized in the Figure \ref{visadj}, "EV" and "N2V" are more dense and more sparse than STWave respectively, which demonstrate the equilibrium of the graph wavelet. We further visualize the ground truth and the predicted curves of traffic flow in Figure \ref{vis} and Appendix C. The predicted curves of the stable component (the red rectangle) of STWave are more accurate than that of STGODE because the supervision of the low-frequency component benefits the prediction of long-term trends. In particular, STWave significantly outperforms STGODE for the fluctuation time points (the green rectangle) because we can obtain and learn the high-frequency pattern individually by the discrete wavelet transform.
\section{Conclusion}
In this paper, we propose a novel STWave for traffic flow forecasting. Specifically, STWave is equipped with a DWT to disentangle intricate traffic sequences, whereby a dual-channel encoder is proposed, \emph{i.e.}, dilated causal convolution and temporal attention, to better represent the different dependencies. Furthermore, with the efficient spectral graph attention network, STWave refines the spatial representation under the global context efficiently and effectively. The frequency-specific decoder then merges and exploits the latent representations of the low- and high-frequency components by fusion attention and multi-supervision. Experimental results show the superiority of STWave over baselines.
\bibliographystyle{named}
\bibliography{ijcai22}

\begin{thebibliography}{}

\bibitem[\protect\citeauthoryear{Bai \bgroup \em et al.\egroup
  }{2020}]{DBLP:conf/nips/0001YL0020}
Lei Bai, Lina Yao, Can Li, Xianzhi Wang, and Can Wang.
\newblock Adaptive graph convolutional recurrent network for traffic
  forecasting.
\newblock In {\em Proceedings of NeurIPS}, 2020.

\bibitem[\protect\citeauthoryear{Berndt and Clifford}{1994}]{berndt1994using}
Donald~J Berndt and James Clifford.
\newblock Using dynamic time warping to find patterns in time series.
\newblock In {\em Proceedings of KDD workshop}, 1994.

\bibitem[\protect\citeauthoryear{Daubechies}{1992}]{daubechies1992ten}
Ingrid Daubechies.
\newblock {\em Ten lectures on wavelets}.
\newblock SIAM, 1992.

\bibitem[\protect\citeauthoryear{Defferrard \bgroup \em et al.\egroup
  }{2016}]{defferrard2016convolutional}
Micha{\"e}l Defferrard, Xavier Bresson, and Pierre Vandergheynst.
\newblock Convolutional neural networks on graphs with fast localized spectral
  filtering.
\newblock In {\em Proceedings of Neurips}, 2016.

\bibitem[\protect\citeauthoryear{Dwivedi and
  Bresson}{2020}]{dwivedi2020generalization}
Vijay~Prakash Dwivedi and Xavier Bresson.
\newblock A generalization of transformer networks to graphs.
\newblock {\em arXiv preprint arXiv:2012.09699}, 2020.

\bibitem[\protect\citeauthoryear{Elmi}{2020}]{elmi2020deep}
Sayda Elmi.
\newblock Deep stacked residual neural network and bidirectional lstm for speed
  prediction on real-life traffic data.
\newblock In {\em Proceedings of ECAI}, 2020.

\bibitem[\protect\citeauthoryear{Fang \bgroup \em et al.\egroup
  }{2021}]{fang2021spatial}
Zheng Fang, Qingqing Long, Guojie Song, and Kunqing Xie.
\newblock Spatial-temporal graph ode networks for traffic flow forecasting.
\newblock In {\em Proceedings of SIGKDD}, 2021.

\bibitem[\protect\citeauthoryear{Grover and
  Leskovec}{2016}]{grover2016node2vec}
Aditya Grover and Jure Leskovec.
\newblock node2vec: Scalable feature learning for networks.
\newblock In {\em Proceedings of SIGKDD}, 2016.

\bibitem[\protect\citeauthoryear{Guo \bgroup \em et al.\egroup
  }{2019}]{guo2019attention}
Shengnan Guo, Youfang Lin, Ning Feng, Chao Song, and Huaiyu Wan.
\newblock Attention based spatial-temporal graph convolutional networks for
  traffic flow forecasting.
\newblock In {\em Proceedings of AAAI}, 2019.

\bibitem[\protect\citeauthoryear{Guo \bgroup \em et al.\egroup
  }{2021}]{guo2021hierarchical}
Kan Guo, Yongli Hu, Yanfeng Sun, Sean Qian, Junbin Gao, and Baocai Yin.
\newblock Hierarchical graph convolution networks for traffic forecasting.
\newblock In {\em Proceedings of AAAI}, 2021.

\bibitem[\protect\citeauthoryear{Hamilton}{2020}]{hamilton2020time}
James~Douglas Hamilton.
\newblock {\em Time series analysis}.
\newblock Princeton university press, 2020.

\bibitem[\protect\citeauthoryear{Huang \bgroup \em et al.\egroup
  }{2020}]{huang2020lsgcn}
Rongzhou Huang, Chuyin Huang, Yubao Liu, Genan Dai, and Weiyang Kong.
\newblock Lsgcn: Long short-term traffic prediction with graph convolutional
  networks.
\newblock In {\em Proceedings of IJCAI}, 2020.

\bibitem[\protect\citeauthoryear{Li and Zhu}{2021}]{DBLP:conf/aaai/LiZ21}
Mengzhang Li and Zhanxing Zhu.
\newblock Spatial-temporal fusion graph neural networks for traffic flow
  forecasting.
\newblock In {\em Proceedings of AAAI}, 2021.

\bibitem[\protect\citeauthoryear{Li \bgroup \em et al.\egroup
  }{2018}]{DBLP:conf/iclr/LiYS018}
Yaguang Li, Rose Yu, Cyrus Shahabi, and Yan Liu.
\newblock Diffusion convolutional recurrent neural network: Data-driven traffic
  forecasting.
\newblock In {\em Proceedings of ICLR}, 2018.

\bibitem[\protect\citeauthoryear{Lu \bgroup \em et al.\egroup
  }{2016}]{lu2016integrating}
Zheng Lu, Chen Zhou, Jing Wu, Hao Jiang, and Songyue Cui.
\newblock Integrating granger causality and vector auto-regression for traffic
  prediction of large-scale wlans.
\newblock {\em KSII Transactions on Internet and Information Systems (TIIS)},
  10(1):136--151, 2016.

\bibitem[\protect\citeauthoryear{Lv \bgroup \em et al.\egroup
  }{2018}]{lv2018lc}
Zhongjian Lv, Jiajie Xu, Kai Zheng, Hongzhi Yin, Pengpeng Zhao, and Xiaofang
  Zhou.
\newblock Lc-rnn: A deep learning model for traffic speed prediction.
\newblock In {\em Proceedings of IJCAI}, 2018.

\bibitem[\protect\citeauthoryear{Park \bgroup \em et al.\egroup
  }{2020}]{park2020st}
Cheonbok Park, Chunggi Lee, Hyojin Bahng, Yunwon Tae, Seungmin Jin, Kihwan Kim,
  Sungahn Ko, and Jaegul Choo.
\newblock St-grat: A novel spatio-temporal graph attention networks for
  accurately forecasting dynamically changing road speed.
\newblock In {\em Proceedings of CIKM}, 2020.

\bibitem[\protect\citeauthoryear{Song \bgroup \em et al.\egroup
  }{2020}]{song2020spatial}
Chao Song, Youfang Lin, Shengnan Guo, and Huaiyu Wan.
\newblock Spatial-temporal synchronous graph convolutional networks: A new
  framework for spatial-temporal network data forecasting.
\newblock In {\em Proceedings of AAAI}, 2020.

\bibitem[\protect\citeauthoryear{Sutskever \bgroup \em et al.\egroup
  }{2014}]{sutskever2014sequence}
Ilya Sutskever, Oriol Vinyals, and Quoc~V Le.
\newblock Sequence to sequence learning with neural networks.
\newblock In {\em Proceedings of NeurIPS}, 2014.

\bibitem[\protect\citeauthoryear{Van~Lint and
  Van~Hinsbergen}{2012}]{van2012short}
JWC Van~Lint and CPIJ Van~Hinsbergen.
\newblock Short-term traffic and travel time prediction models.
\newblock {\em Artificial Intelligence Applications to Critical Transportation
  Issues}, 22(1):22--41, 2012.

\bibitem[\protect\citeauthoryear{Williams and
  Hoel}{2003}]{williams2003modeling}
Billy~M Williams and Lester~A Hoel.
\newblock Modeling and forecasting vehicular traffic flow as a seasonal arima
  process: Theoretical basis and empirical results.
\newblock {\em Journal of transportation engineering}, 129(6):664--672, 2003.

\bibitem[\protect\citeauthoryear{Wu \bgroup \em et al.\egroup
  }{2004}]{wu2004travel}
Chun-Hsin Wu, Jan-Ming Ho, and Der-Tsai Lee.
\newblock Travel-time prediction with support vector regression.
\newblock {\em IEEE transactions on intelligent transportation systems},
  5(4):276--281, 2004.

\bibitem[\protect\citeauthoryear{Wu \bgroup \em et al.\egroup
  }{2019}]{DBLP:conf/ijcai/WuPLJZ19}
Zonghan Wu, Shirui Pan, Guodong Long, Jing Jiang, and Chengqi Zhang.
\newblock Graph wavenet for deep spatial-temporal graph modeling.
\newblock In {\em Proceedings of IJCAI}, 2019.

\bibitem[\protect\citeauthoryear{Xu \bgroup \em et al.\egroup
  }{2019}]{DBLP:conf/iclr/XuSCQC19}
Bingbing Xu, Huawei Shen, Qi~Cao, Yunqi Qiu, and Xueqi Cheng.
\newblock Graph wavelet neural network.
\newblock In {\em Proceedings of ICLR}, 2019.

\bibitem[\protect\citeauthoryear{Xu \bgroup \em et al.\egroup
  }{2021}]{xu2021autoformer}
Jiehui Xu, Jianmin Wang, Mingsheng Long, et~al.
\newblock Autoformer: Decomposition transformers with auto-correlation for
  long-term series forecasting.
\newblock {\em Proceedings of NeurIPS}, 2021.

\bibitem[\protect\citeauthoryear{Yu \bgroup \em et al.\egroup
  }{2018}]{DBLP:conf/ijcai/YuYZ18}
Bing Yu, Haoteng Yin, and Zhanxing Zhu.
\newblock Spatio-temporal graph convolutional networks: {A} deep learning
  framework for traffic forecasting.
\newblock In {\em Proceedings of IJCAI}, 2018.

\bibitem[\protect\citeauthoryear{Zhang \bgroup \em et al.\egroup
  }{2020}]{zhang2020spatial}
Xiyue Zhang, Chao Huang, Yong Xu, and Lianghao Xia.
\newblock Spatial-temporal convolutional graph attention networks for citywide
  traffic flow forecasting.
\newblock In {\em Proceedings of CIKM}, 2020.

\bibitem[\protect\citeauthoryear{Zheng \bgroup \em et al.\egroup
  }{2020}]{zheng2020gman}
Chuanpan Zheng, Xiaoliang Fan, Cheng Wang, and Jianzhong Qi.
\newblock Gman: A graph multi-attention network for traffic prediction.
\newblock In {\em Proceedings of AAAI}, 2020.

\bibitem[\protect\citeauthoryear{Zhou \bgroup \em et al.\egroup
  }{2021}]{zhou2021informer}
Haoyi Zhou, Shanghang Zhang, Jieqi Peng, Shuai Zhang, Jianxin Li, Hui Xiong,
  and Wancai Zhang.
\newblock Informer: Beyond efficient transformer for long sequence time-series
  forecasting.
\newblock In {\em Proceedings of AAAI}, 2021.

\end{thebibliography}
\newpage
\appendix
\section{Related Work}
\paragraph{Traffic Forecasting.} Early researches use the traditional statistical methods to predict traffic flow \cite{hamilton2020time,williams2003modeling}, yet they base on linear assumptions, which fail to capture the non-linear dependencies in the traffic flow forecasting. \cite{wu2004travel,van2012short} apply machine learning methods in traffic flow forecasting, but the hand-craft features limit their ability to generalize. With the success of deep learning in computer vision and natural language processing, a line of research models temporal patterns for each road individually, such as LSTM \cite{elmi2020deep}, TCN \cite{sutskever2014sequence}, and Transformer \cite{zhou2021informer}. Another line combines GCNs with sequential methods to capture spatio-temporal patterns simultaneously, such as STGCN \cite{DBLP:conf/ijcai/YuYZ18} and DCRNN \cite{DBLP:conf/iclr/LiYS018}. Graph WaveNet \cite{DBLP:conf/ijcai/WuPLJZ19} and AGCRN \cite{DBLP:conf/nips/0001YL0020} further combine adaptive graph convolution with TCN and RNN to capture spatio-temporal dependencies through back-propagation. STFGNN \cite{DBLP:conf/aaai/LiZ21} expands the spatial receptive field by a novel fusion operation of various spatial and temporal graphs. STGODE \cite{fang2021spatial} utilizes the tensor-based ODE to increase the depth of GCN. However, GCN-based methods fail to capture dynamic spatial patterns.
\paragraph{Graph Attention Network for Traffic Forecasting.} ST-CGA \cite{zhang2020spatial} proposes a graph attention network (GAT) based method to learn the weights between neighbor roads in each time step. ASTGCN \cite{guo2019attention} further utilizes the attention mechanism on spatio-temporal dimensions to adjust the weights of spatio-temporal convolution. LSGCN \cite{huang2020lsgcn} integrates a novel graph attention network and graph convolution into a spatial gated block to capture spatial dependence. ST-GRAT \cite{park2020st} drops the input graph in the vanilla GAT to derive the full GAT. The full GAT can alleviate the influence of hard inductive bias and capture the global spatial dependence. Similar to ST-GRAT, GMAN \cite{zheng2020gman} utilizes the full GAT with a graph positional encoding derived by node2vec \cite{grover2016node2vec} to bring structural information into model. Compared with the graph positional encoding that takes advantage of the graph wavelet in STWave, node2vec is inefficient to extract global information and brings extra parameters. Moreover, the complexity of full GAT-based methods is $O(N^2)$.
\section{Prediction for Each Time Step}
\begin{figure*}[t]
    \centering
        \begin{subfigure}{0.33\linewidth}
        \includegraphics[width=\linewidth]{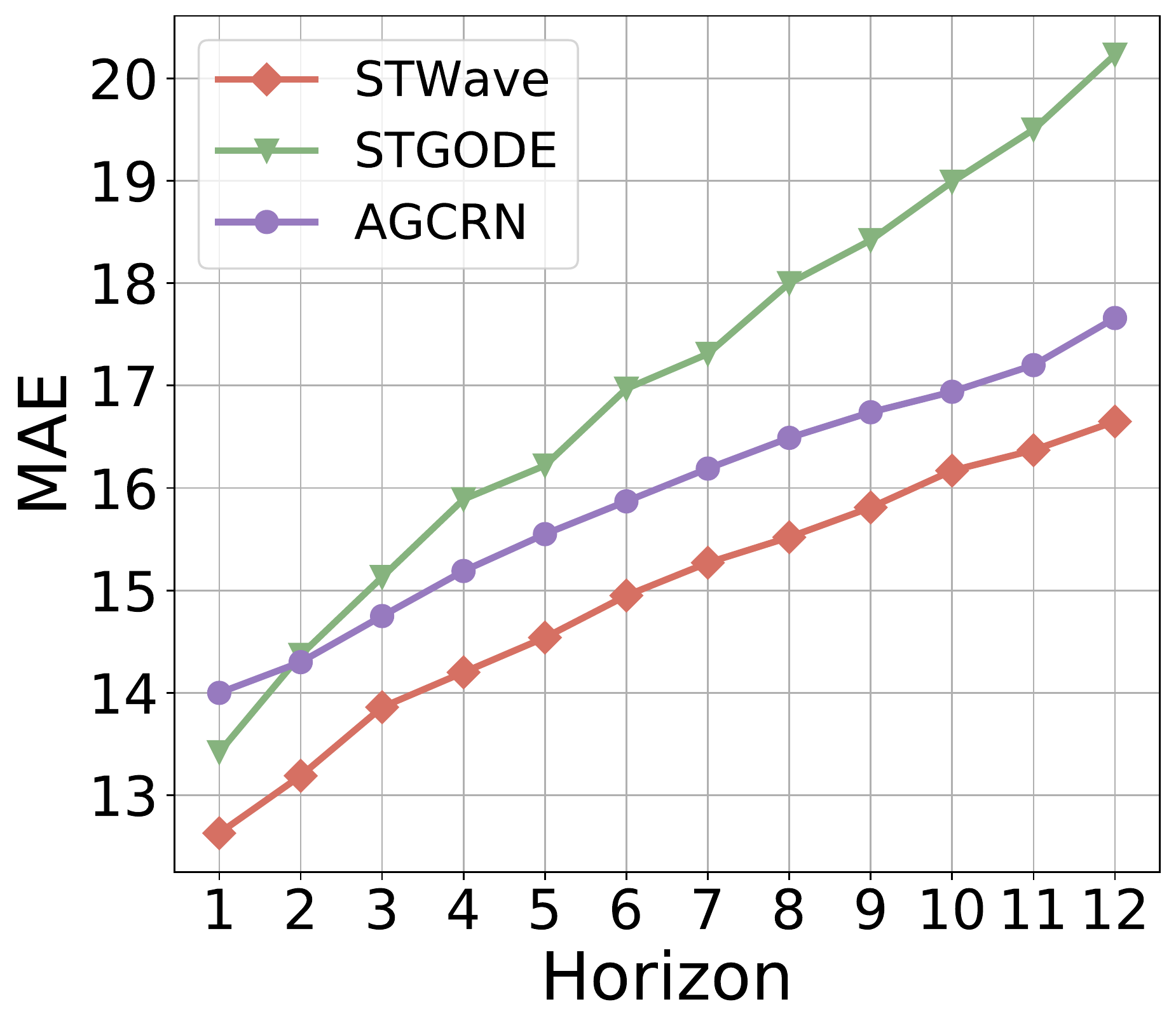}
        \captionsetup{font=small}
        \caption{MAE on PeMSD3.}
      \end{subfigure}%
      \hfill
      \begin{subfigure}{0.33\linewidth}
        \includegraphics[width=\linewidth]{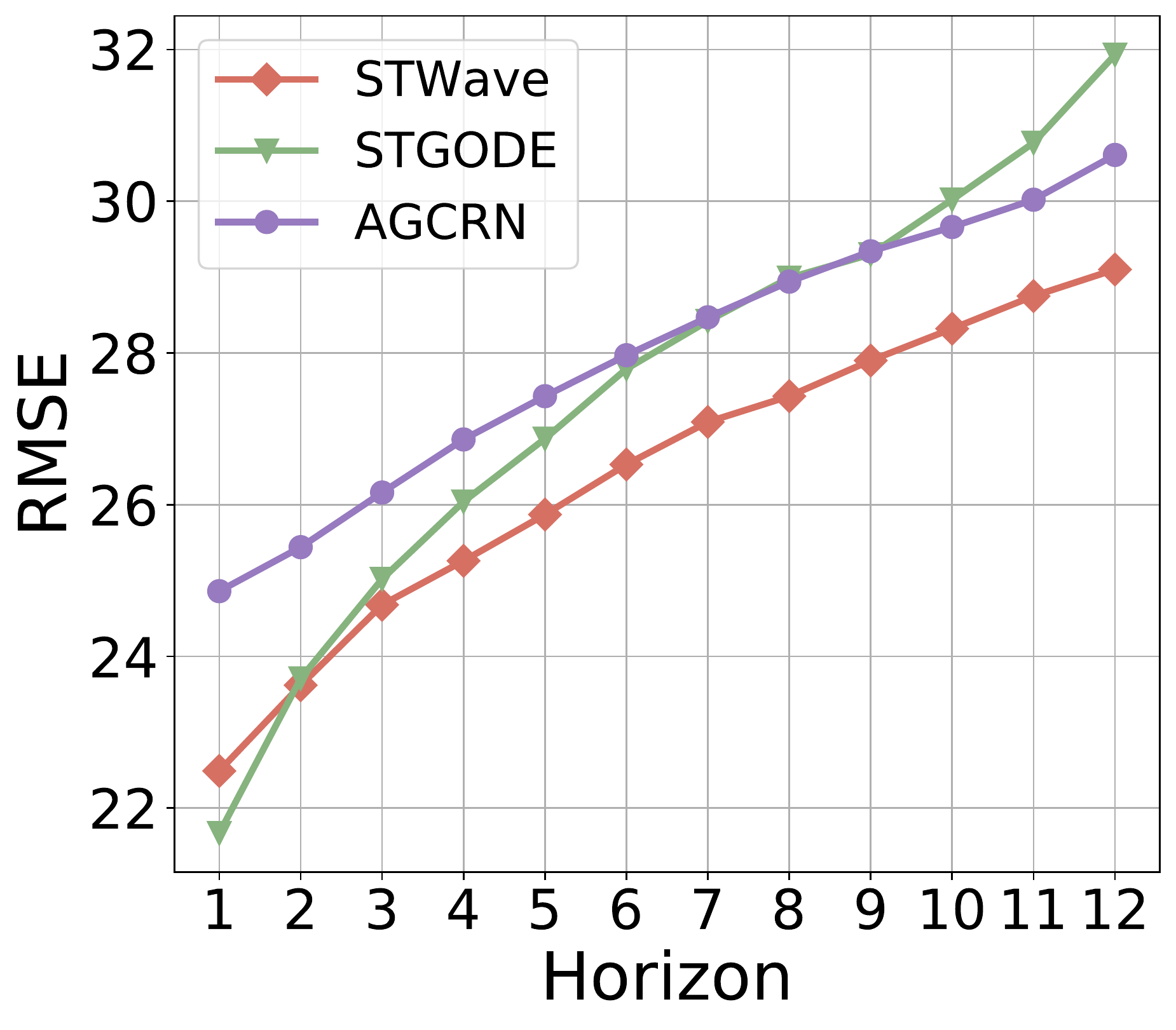}
        \captionsetup{font=small}
        \caption{RMSE on PeMSD3.}
      \end{subfigure}
      \hfill
      \begin{subfigure}{0.33\linewidth}
        \includegraphics[width=\linewidth]{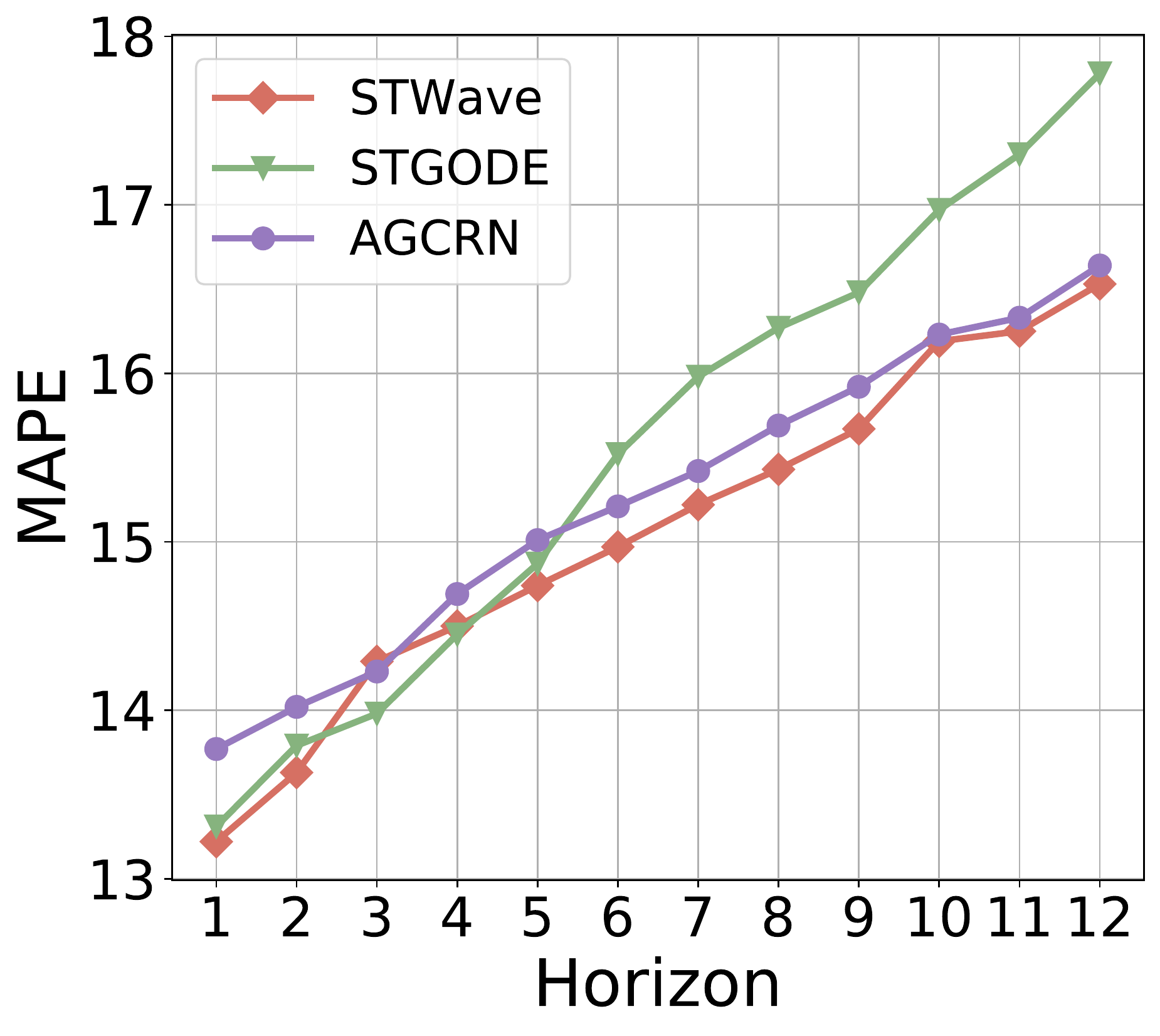}
        \captionsetup{font=small}
        \caption{MAPE on PeMSD3.}
      \end{subfigure}
      
    \begin{subfigure}{0.33\linewidth}
        \includegraphics[width=\linewidth]{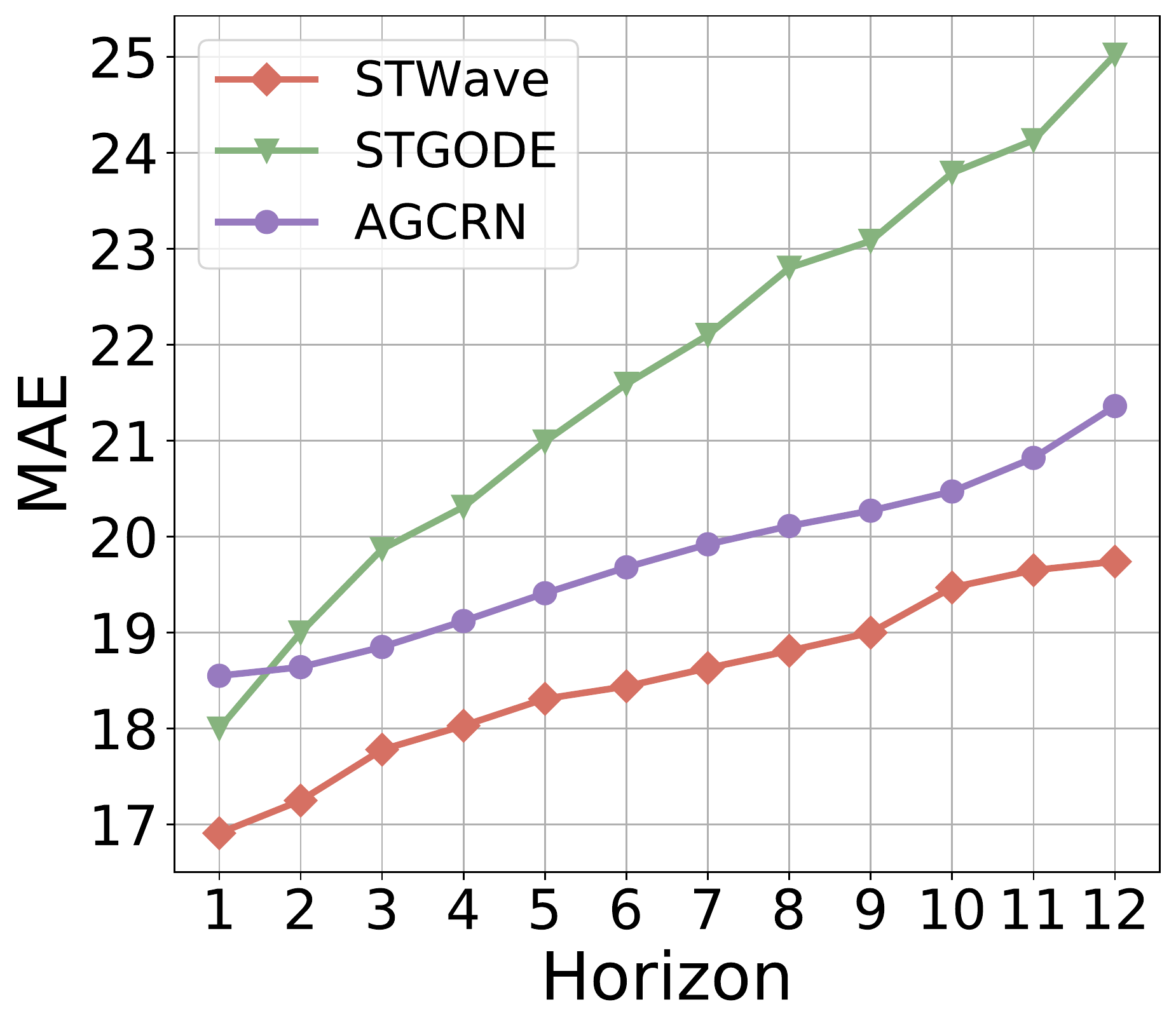}
        \captionsetup{font=small}
        \caption{MAE on PeMSD4.}
      \end{subfigure}%
      \hfill
      \begin{subfigure}{0.33\linewidth}
        \includegraphics[width=\linewidth]{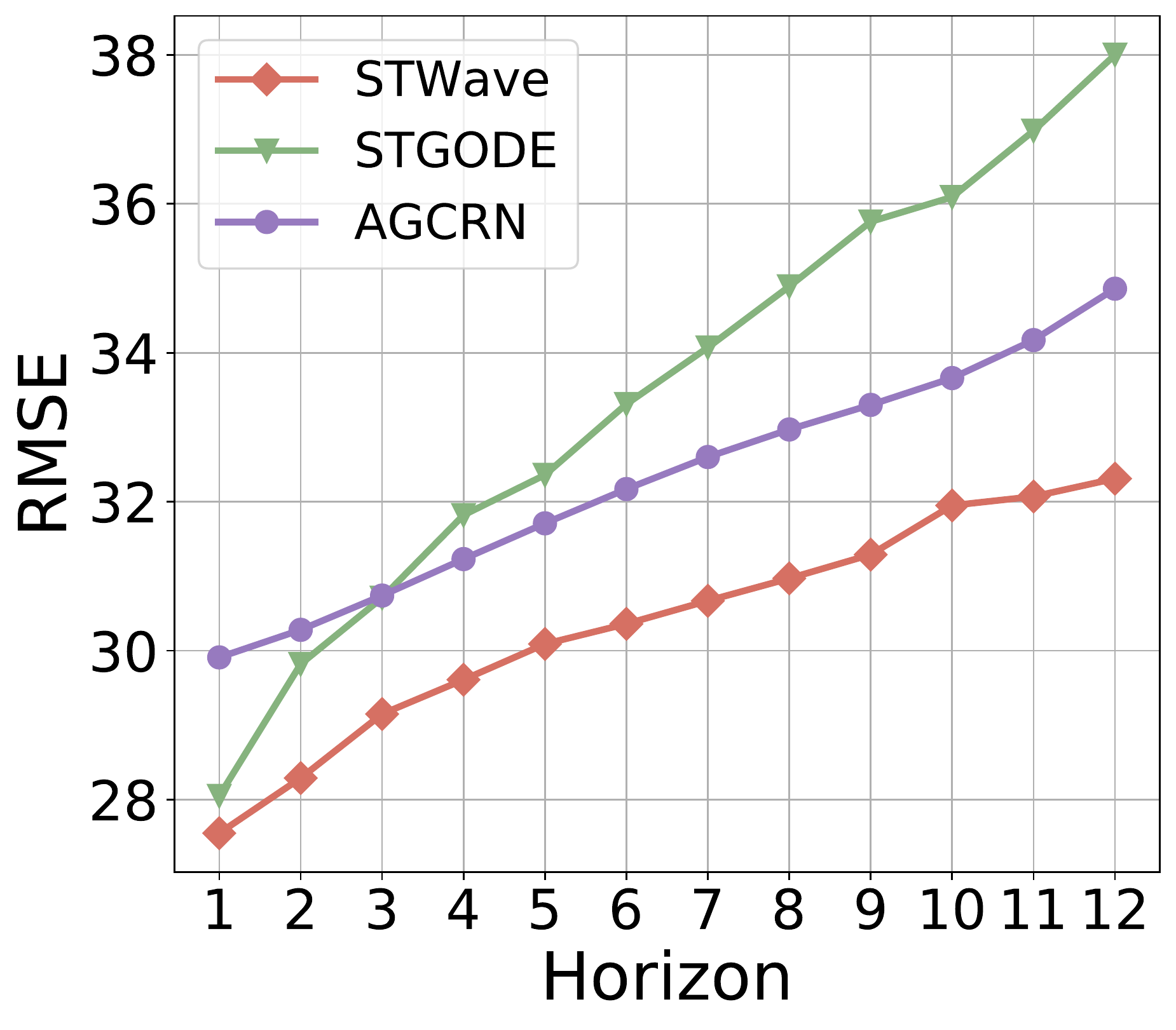}
        \captionsetup{font=small}
        \caption{RMSE on PeMSD4.}
      \end{subfigure}
      \hfill
      \begin{subfigure}{0.33\linewidth}
        \includegraphics[width=\linewidth]{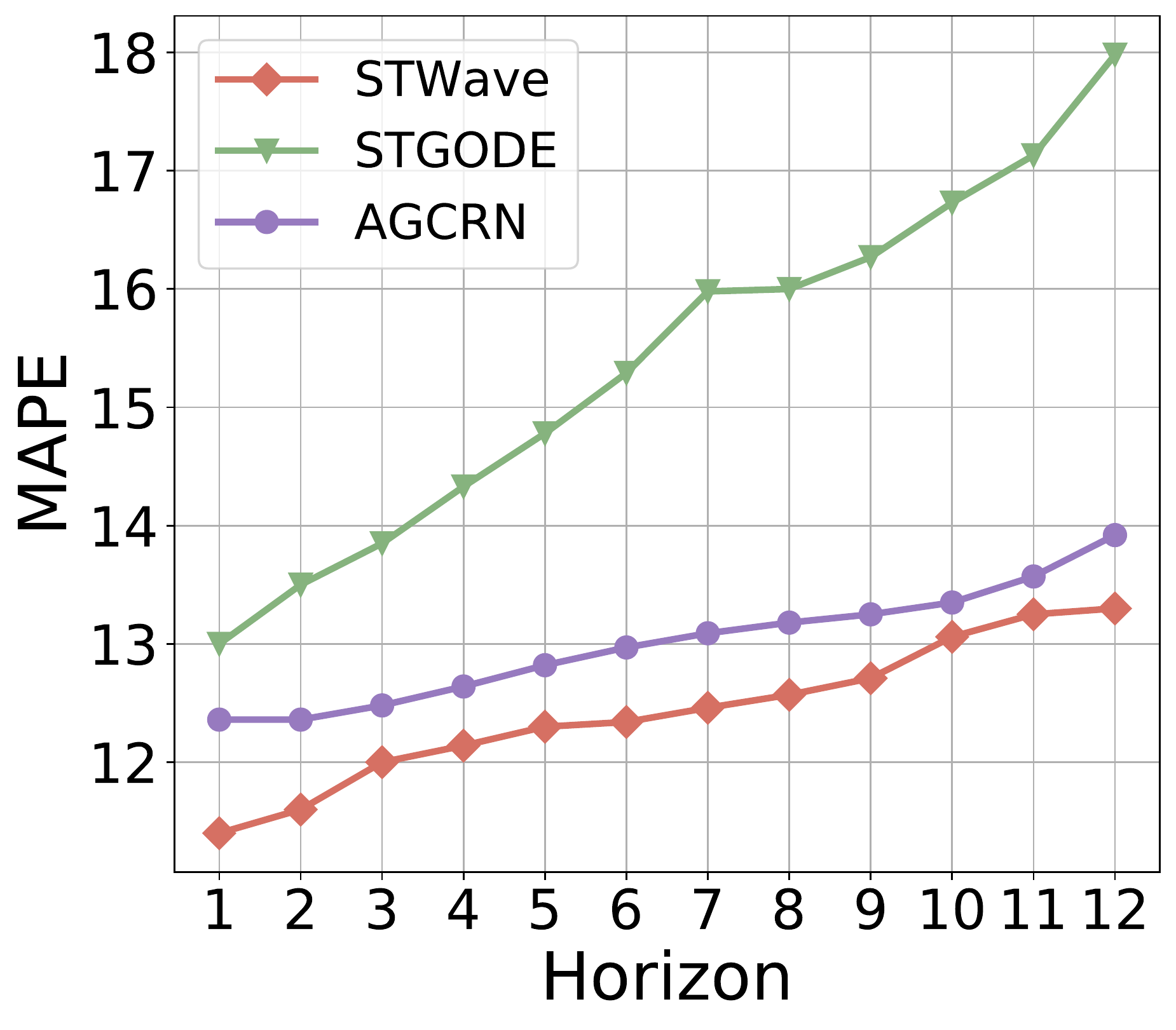}
        \captionsetup{font=small}
        \caption{MAPE on PeMSD4.}
      \end{subfigure}
      
    \begin{subfigure}{0.33\linewidth}
        \includegraphics[width=\linewidth]{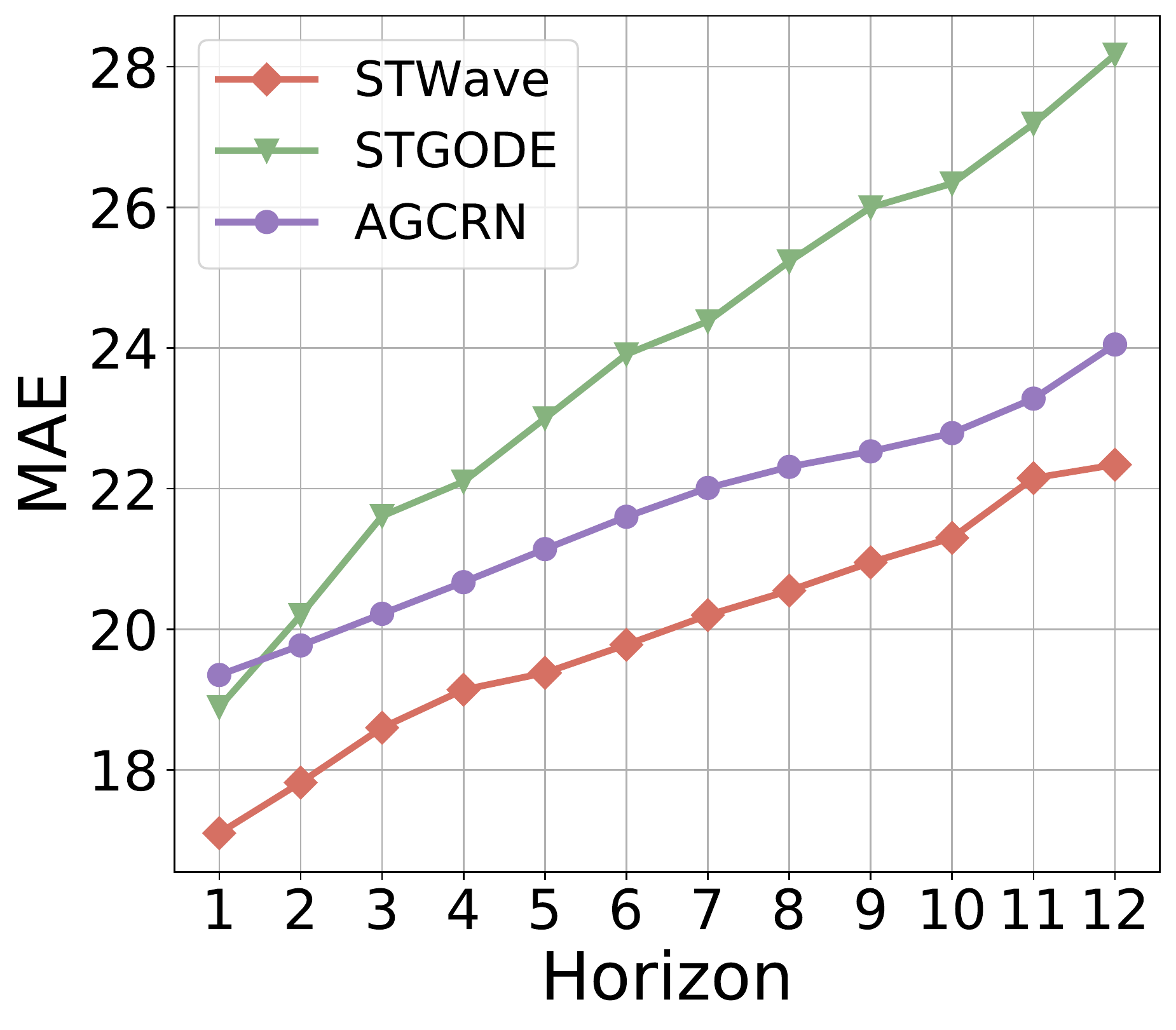}
        \captionsetup{font=small}
        \caption{MAE on PeMSD7.}
      \end{subfigure}%
      \hfill
      \begin{subfigure}{0.33\linewidth}
        \includegraphics[width=\linewidth]{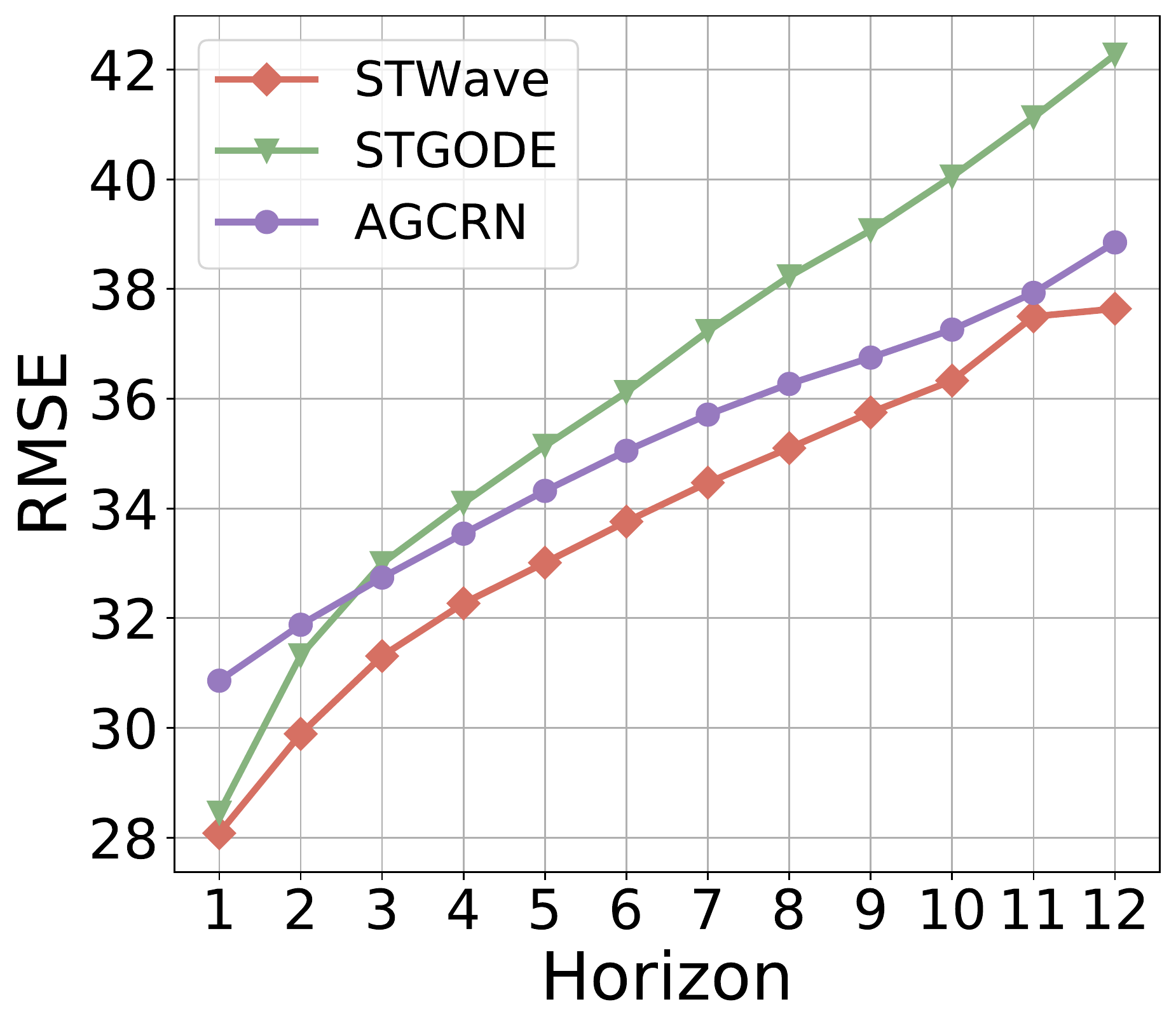}
        \captionsetup{font=small}
        \caption{RMSE on PeMSD7.}
      \end{subfigure}
      \hfill
      \begin{subfigure}{0.33\linewidth}
        \includegraphics[width=\linewidth]{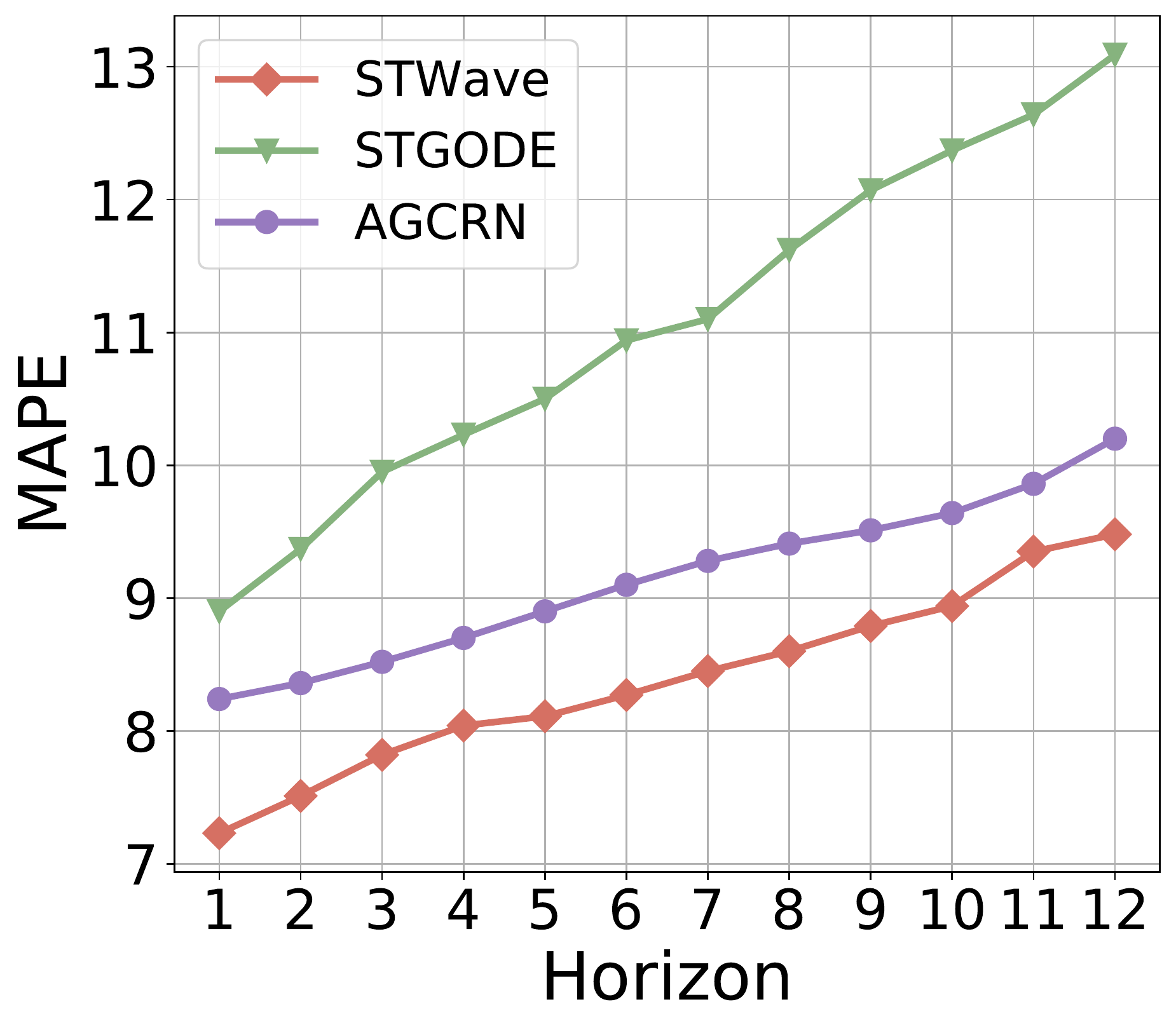}
        \captionsetup{font=small}
        \caption{MAPE on PeMSD7.}
      \end{subfigure}
      
    \begin{subfigure}{0.33\linewidth}
        \includegraphics[width=\linewidth]{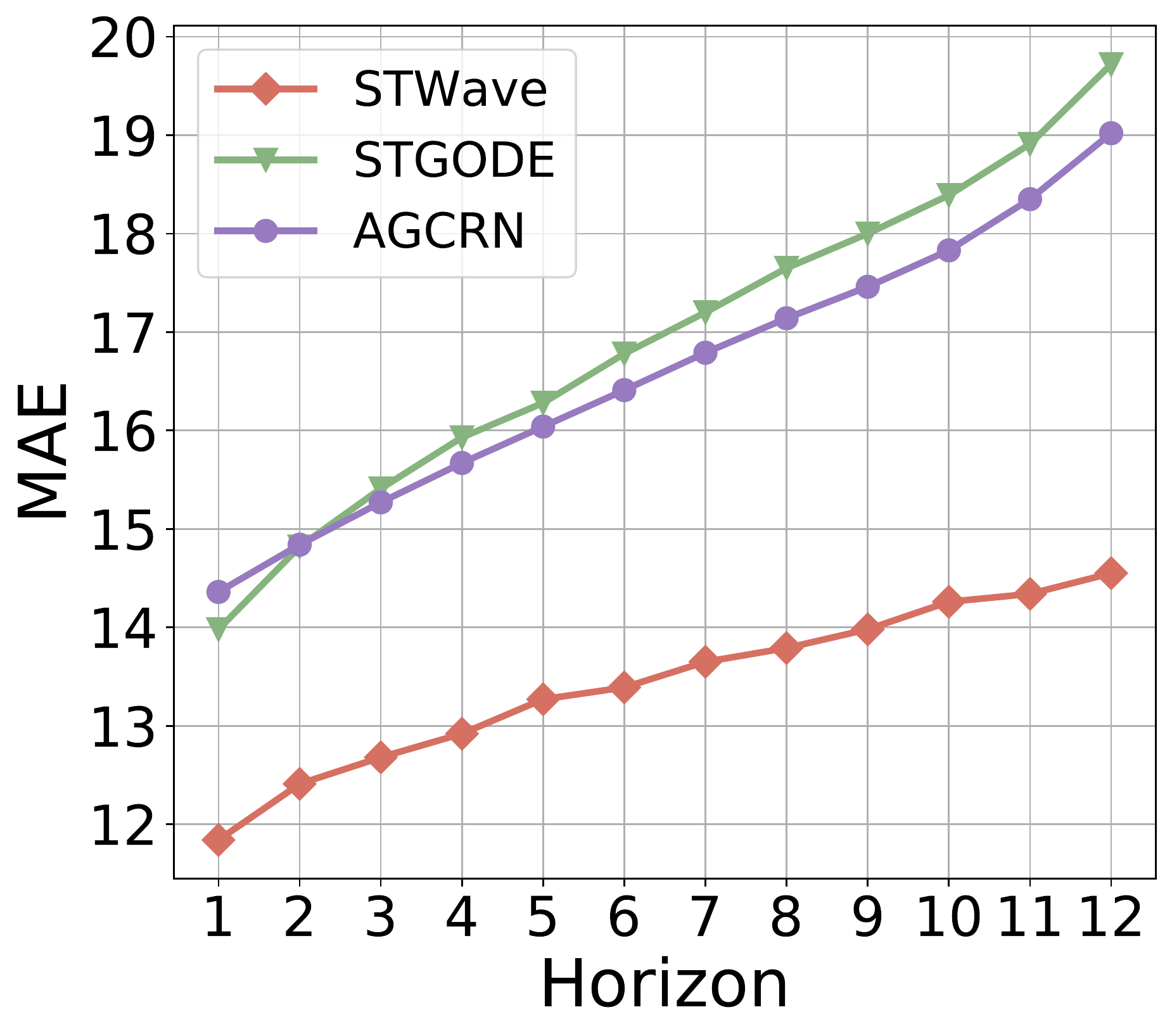}
        \captionsetup{font=small}
        \caption{MAE on PeMSD8.}
      \end{subfigure}%
      \hfill
      \begin{subfigure}{0.33\linewidth}
        \includegraphics[width=\linewidth]{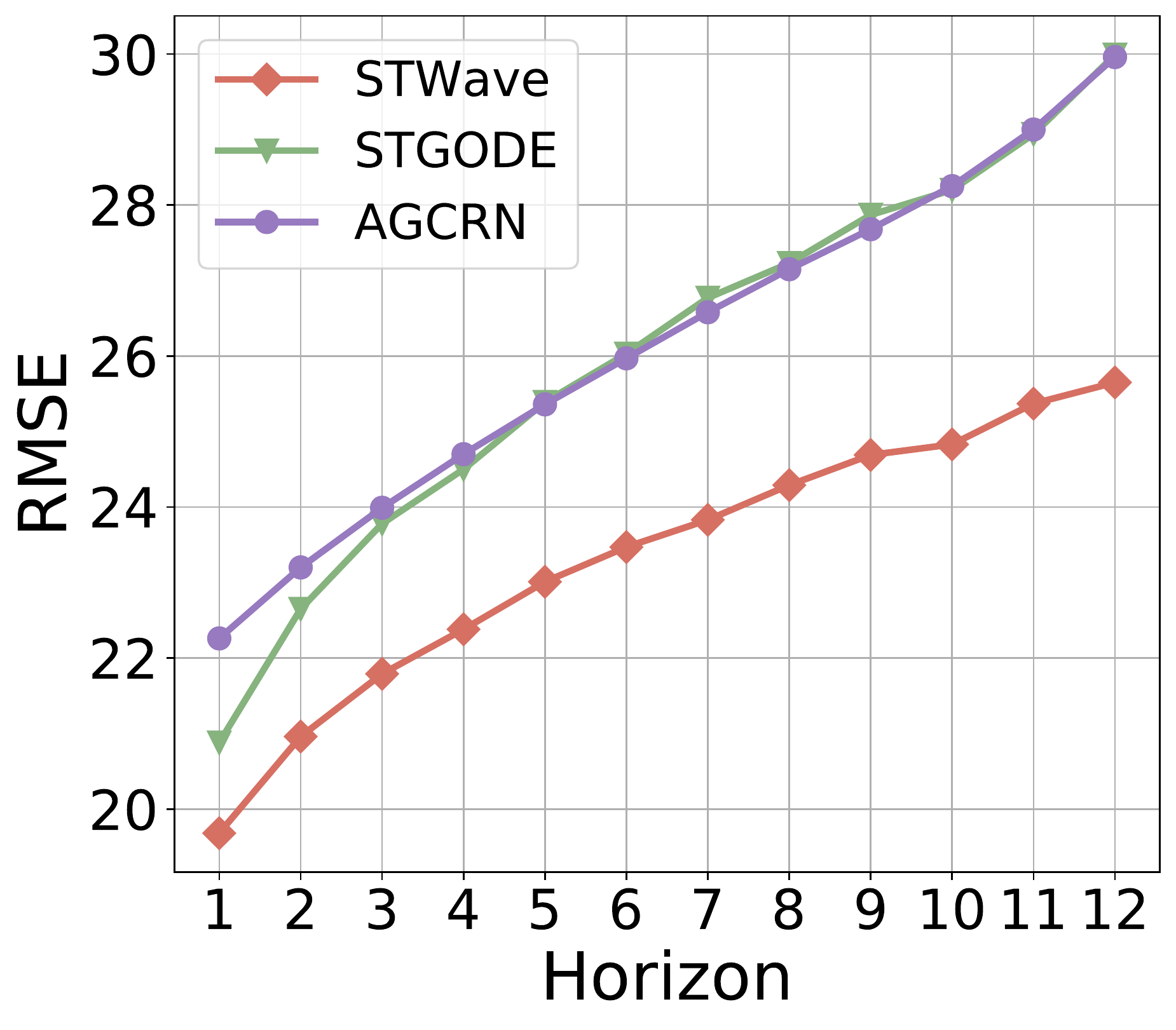}
        \captionsetup{font=small}
        \caption{RMSE on PeMSD8.}
      \end{subfigure}
      \hfill
      \begin{subfigure}{0.33\linewidth}
        \includegraphics[width=\linewidth]{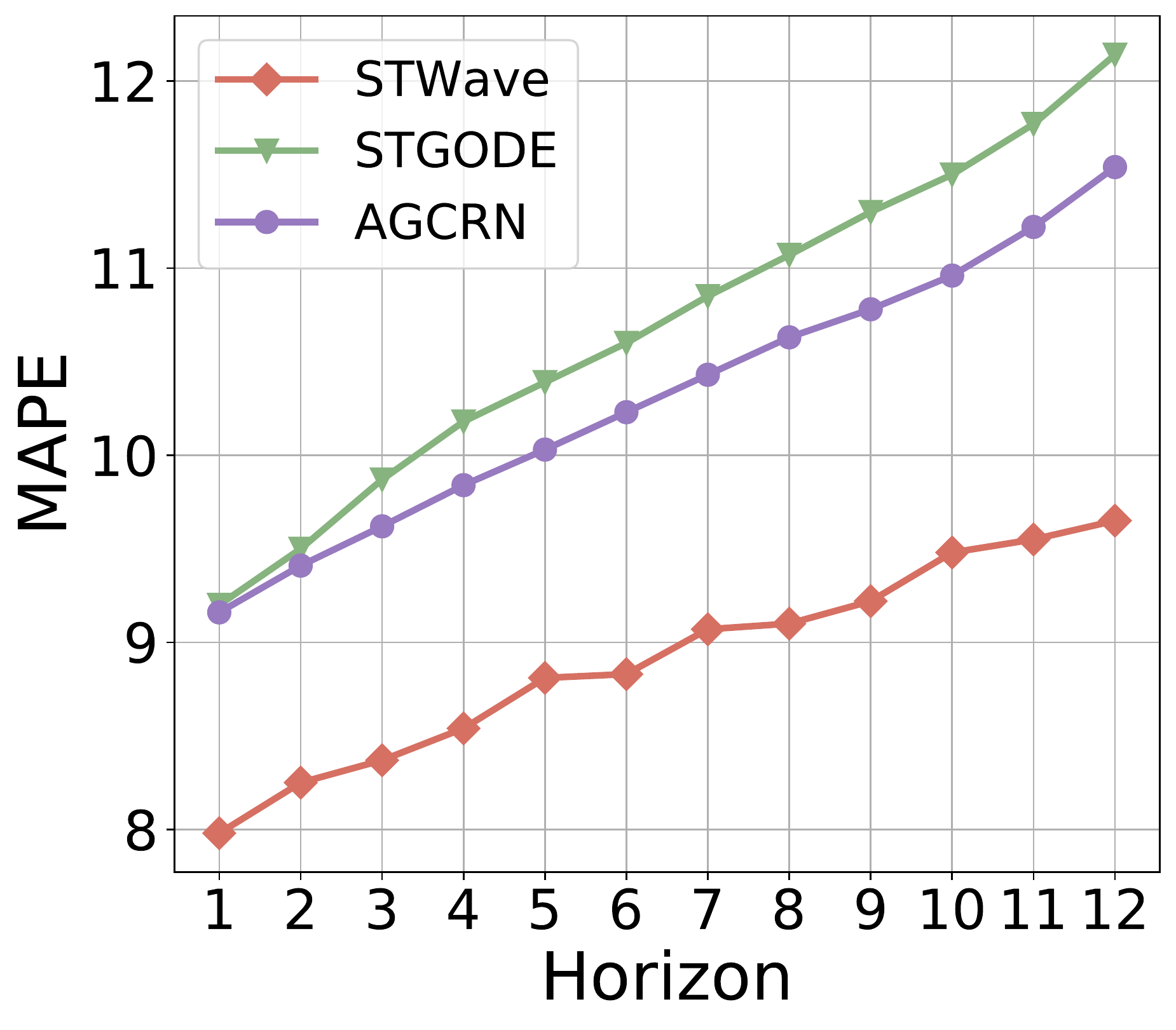}
        \captionsetup{font=small}
        \caption{MAPE on PeMSD8.}
      \end{subfigure}
     
    \caption{Prediction for each time step on all datasets.}
    \label{pre}
\end{figure*}

Figure \ref{pre} shows the prediction error for each time step on all datasets. It is obvious that the error levels show a high correlation to the length in prediction. For all horizons, STWave shows smaller errors than other baselines.
\section{Visualization}
\begin{figure*}[t]
    \centering
        \begin{subfigure}{0.49\linewidth}
        \includegraphics[width=\linewidth]{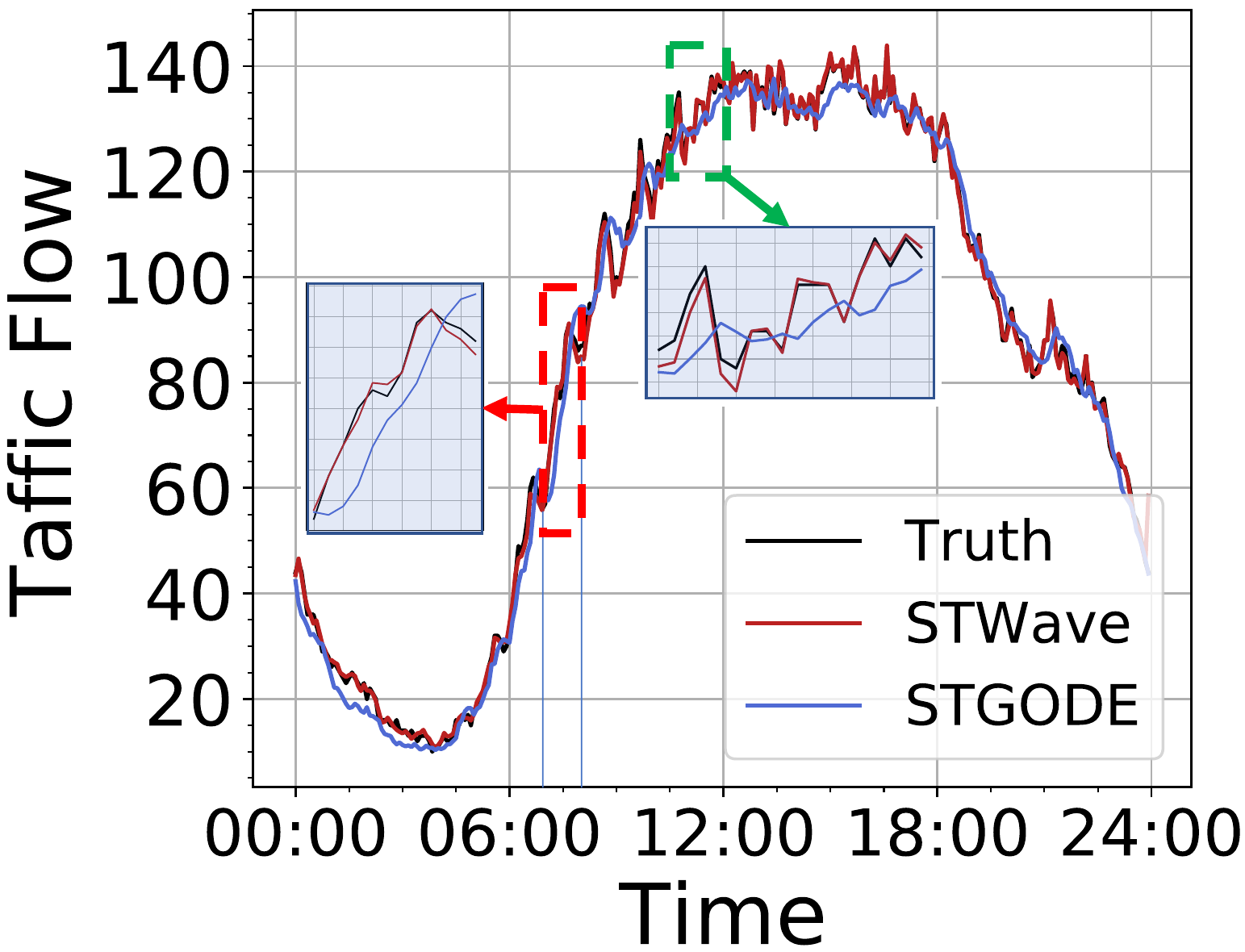}
        \caption{Node 31 in PeMSD3.}
      \end{subfigure}%
      \hfill
      \begin{subfigure}{0.49\linewidth}
        \includegraphics[width=\linewidth]{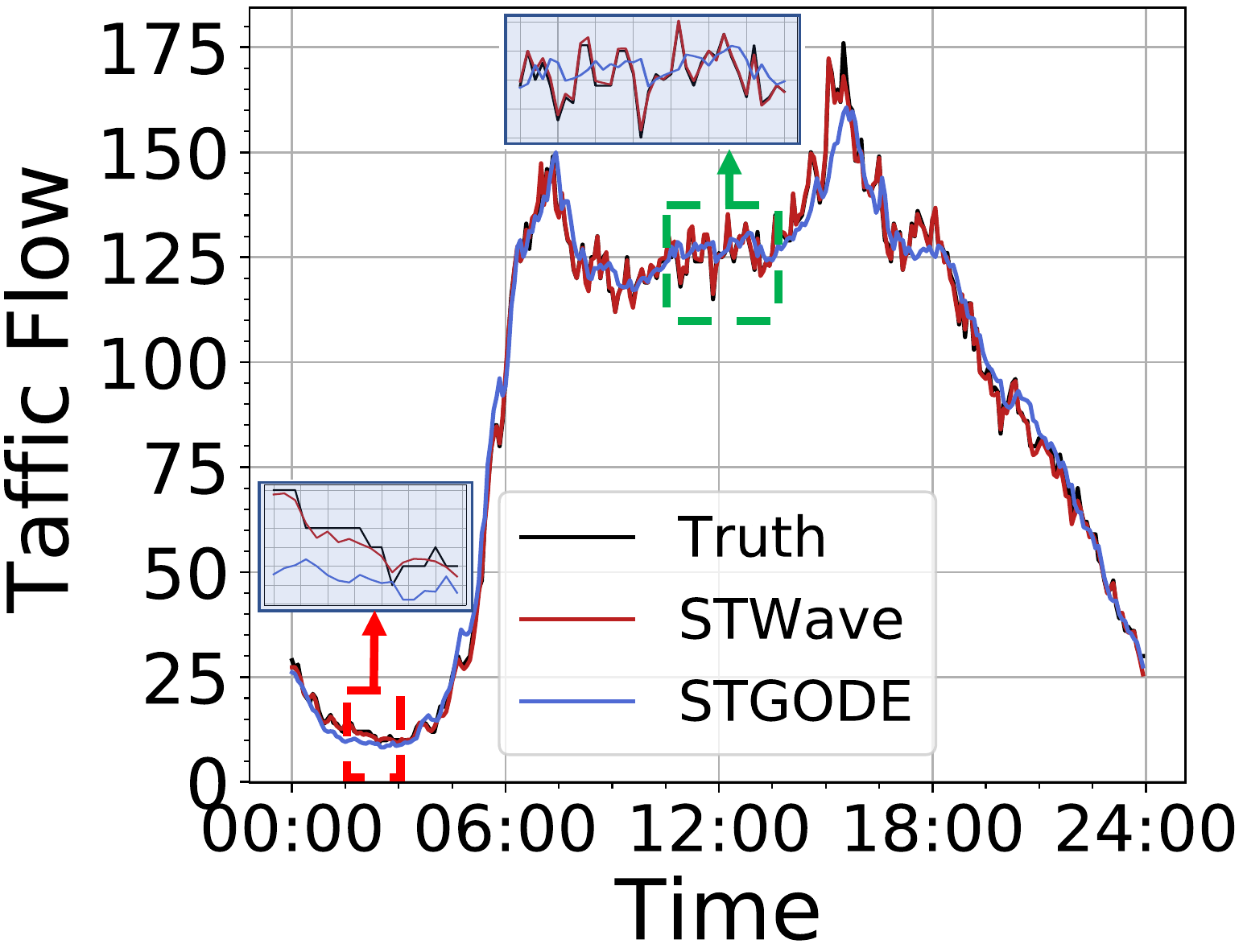}
        \caption{Node 277 in PeMSD3.}
      \end{subfigure}
     
     \begin{subfigure}{0.49\linewidth}
        \includegraphics[width=\linewidth]{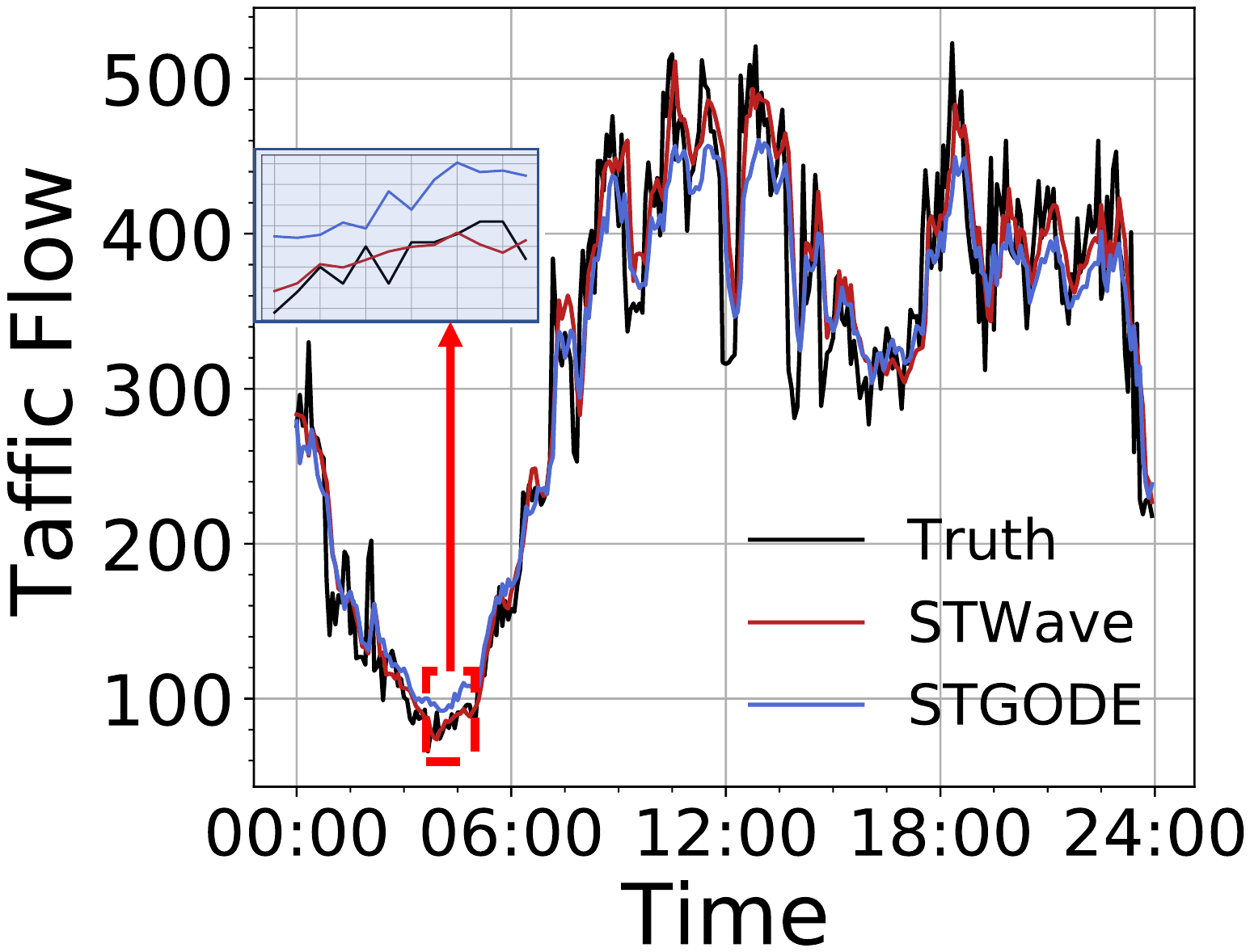}
        \caption{Node 356 in PeMSD7.}
      \end{subfigure}%
      \hfill
      \begin{subfigure}{0.49\linewidth}
        \includegraphics[width=\linewidth]{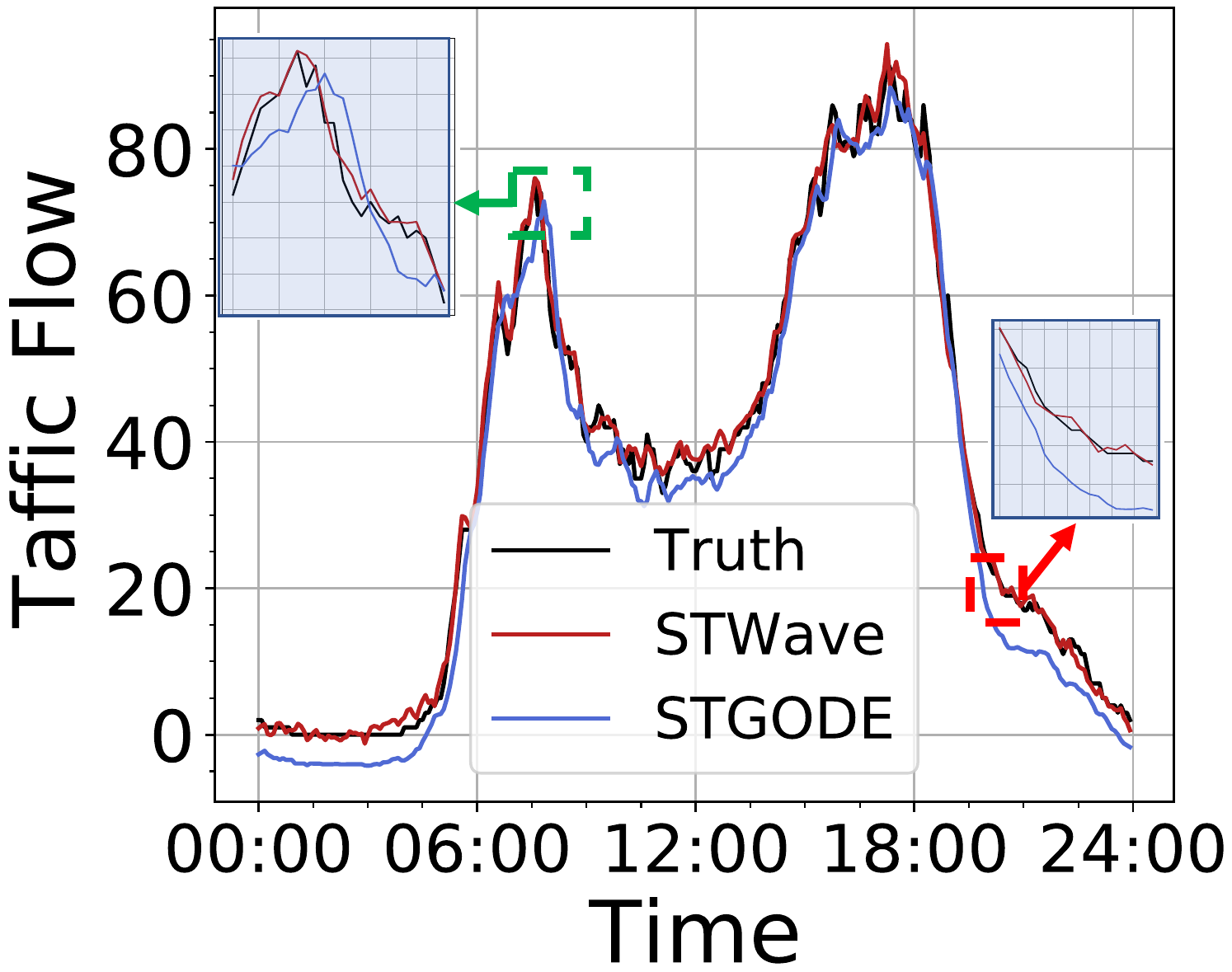}
        \caption{Node 737 in PeMSD7.}
      \end{subfigure}
      
      \begin{subfigure}{0.49\linewidth}
        \includegraphics[width=\linewidth]{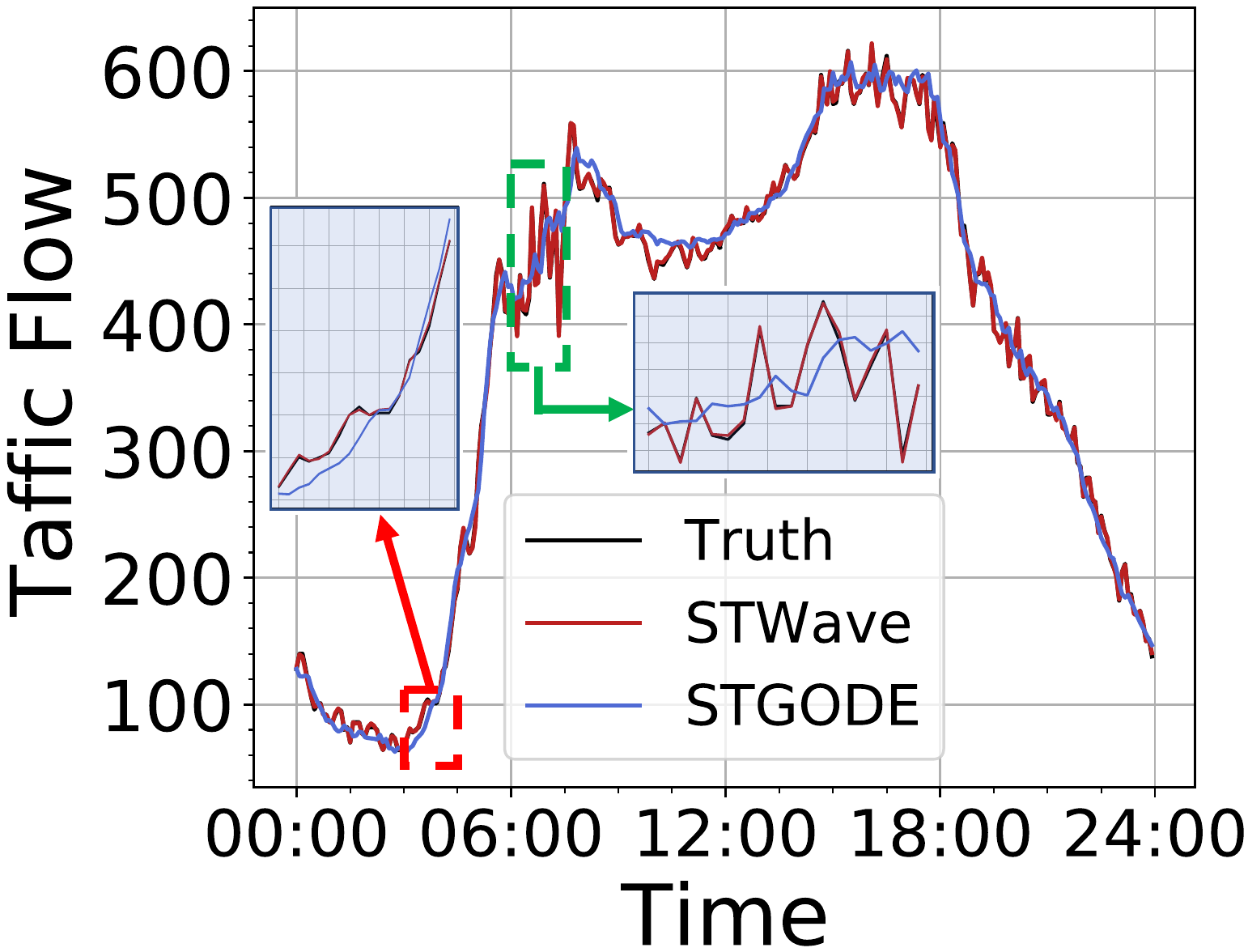}
        \caption{Node 61 in PeMSD8.}
      \end{subfigure}%
      \hfill
      \begin{subfigure}{0.49\linewidth}
        \includegraphics[width=\linewidth]{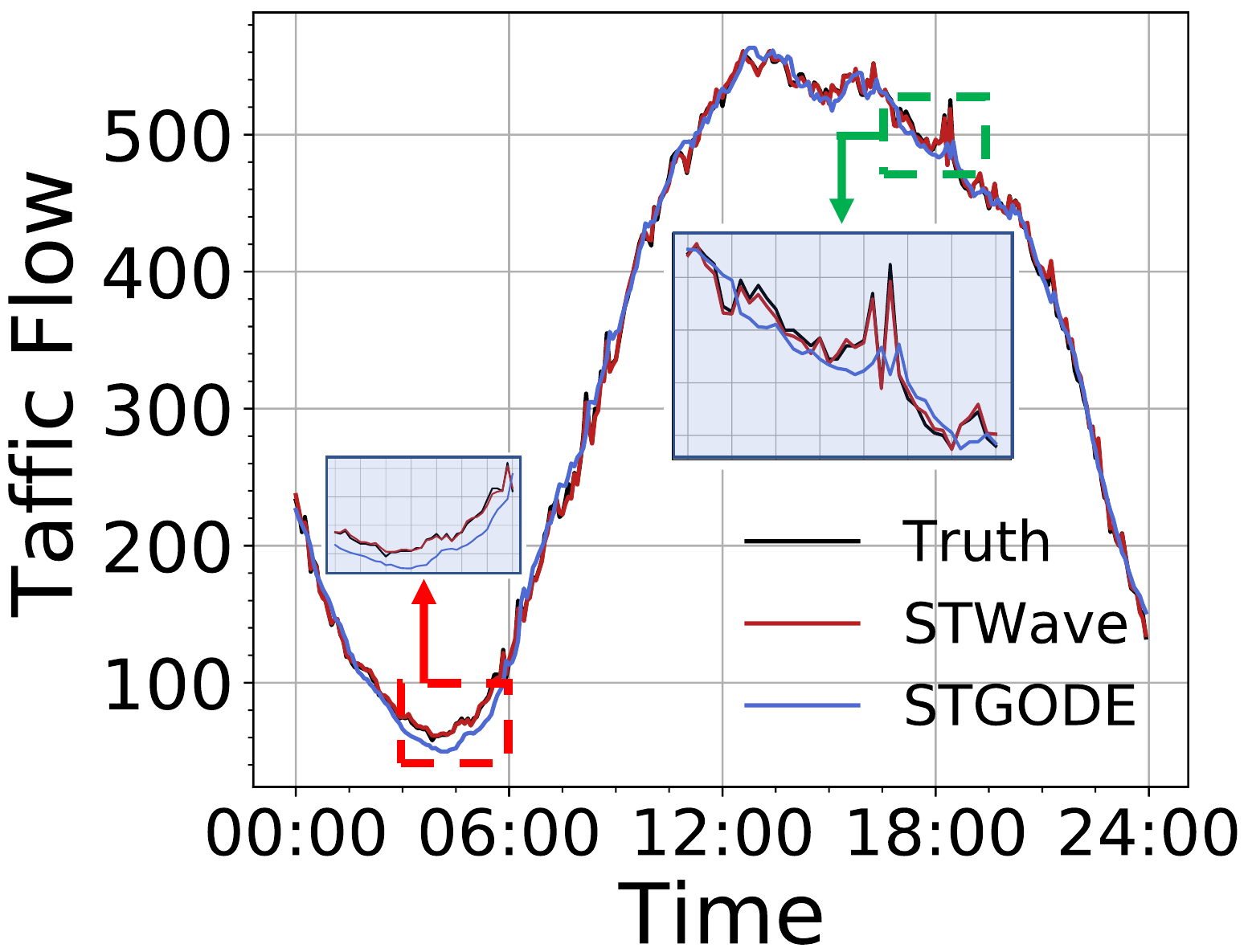}
        \caption{Node 153 in PeMSD8.}
      \end{subfigure}
      
      
      
    \caption{Traffic forecasting visualization.}
    \label{visall}
\end{figure*}

We visualize more examples of the ground truth and the predicted curves of traffic flow in Figure \ref{visall}. Besides, We enlarge the high- and low-frequency components selected by the green and red rectangles respectively.


\end{document}